\documentclass[11pt, a4paper, logo]{googlecloud}

\pdftrailerid{redacted}

\makeatletter
\renewcommand\bibentry[1]{\nocitep{#1}{\frenchspacing\@nameuse{BR@r@#1\@extra@b@citeb}}}
\makeatother

\usepackage{amsmath,amsfonts,bm}

\def\eqref#1{equation~\ref{#1}}

\def\1{\bm{1}}

\DeclareMathAlphabet{\mathsfit}{\encodingdefault}{\sfdefault}{m}{sl}
\SetMathAlphabet{\mathsfit}{bold}{\encodingdefault}{\sfdefault}{bx}{n}

\definecolor{pastelgreen}{rgb}{0.88, 1.0, 0.88}
\definecolor{pastelred}{rgb}{1.0, 0.88, 0.88}
\definecolor{ARCBlue}{RGB}{20, 50, 100}

\usepackage{dsfont}

\usepackage[authoryear, sort&compress, round]{natbib}

\usepackage[utf8]{inputenc} %
\usepackage[T1]{fontenc}    %
\usepackage{siunitx}
\usepackage{microtype}
\usepackage{inconsolata}

\usepackage{booktabs}
\usepackage{graphicx}
\usepackage[most]{tcolorbox}
\usepackage{stackengine}
\usepackage{multirow}
\usepackage{subcaption} 
\usepackage{wrapfig}
\usepackage{hyperref}
\usepackage{makecell}
\usepackage{setspace}
\usepackage{booktabs} 
\usepackage{array}    
\usepackage{geometry} 

\usepackage{url}            
\usepackage{fontawesome5}
\usepackage{amsfonts}       
\usepackage{nicefrac}       
\usepackage{microtype}      
\usepackage{xcolor,colortbl}         
\usepackage{times}
\usepackage{multicol}
\usepackage{blindtext}
\usepackage{tabu}
\usepackage{amsmath, bm}

\usepackage{caption}
\usepackage[normalem]{ulem}
\usepackage{soul}
\usepackage{ragged2e}
\usepackage{pgffor}
\usepackage{textcomp}
\usepackage{amssymb}
\usepackage{pifont}
\usepackage{wrapfig}
\usepackage{booktabs}
\usepackage{tabularx}
\usepackage{algorithm}
\usepackage{algpseudocode}
\usepackage{geometry}
\usepackage{booktabs} 
\usepackage{multirow} 
\usepackage{graphicx} 
\usepackage{paracol}
\usepackage{tcolorbox}
\usepackage{hyperref}
\tcbuselibrary{breakable}
\usepackage[rightcaption]{sidecap}
\usepackage{etoolbox}

\title{HEART: Emotionally-Driven Test-Time Scaling of Language Models}

\correspondingauthor{Gabriela Pinto \href{mailto:gpinto@usc.edu}{<gpinto@usc.edu>}, Palash Goyal \href{mailto:palashgoyal@agoogle.com}{<palashgoyal@google.com>}, and Jinsung Yoon \href{mailto:jinsungyoon@google.com}{<jinsungyoon@google.com>}}

\renewcommand{\today}{}
\author[1,2]{\fontsize{10.0pt}{10.0pt}\selectfont Gabriela Pinto}

\author[1]{\fontsize{10.0pt}{10.0pt}\selectfont Palash Goyal}

\author[1]{\fontsize{10.0pt}{10.0pt}\selectfont Mihir Parmar}

\author[1]{\fontsize{10.0pt}{10.0pt}\selectfont Yiwen Song}

\author[1]{\fontsize{10.0pt}
{10.0pt}\selectfont Souradip Chakraborty}

\author[1]{\fontsize{10.0pt}
{10.0pt}\selectfont Zifeng Wang}

\author[1]{\fontsize{10.0pt}
{10.0pt}\selectfont Jinsung Yoon}

\author[1]{\fontsize{10.0pt}
{10.0pt}\selectfont Tomas Pfister}

\author[1]{\fontsize{10.0pt}
{10.0pt}\selectfont Hamid Palangi}

\affil[1]{\fontsize{9.0pt}{9.0pt}\selectfont Google}
\affil[2]{\fontsize{9.0pt}{9.0pt}\selectfont University of Southern California}

\begin{abstract}
  Test-time scaling has significantly improved how AI models solve problems, yet current methods often get stuck in repetitive, incorrect patterns of thought.  We introduce HEART, a framework that uses emotional cues to guide the model's focus, much like how feelings contribute to human decision-making. By alternating between critical tones to sharpen error detection and encouraging tones to spark new ideas, HEART helps the model break out of dead-end reasoning and find the right solution. We evaluate HEART across seven high-difficulty benchmarks--including Humanity's Last Exam, GPQA Diamond, and LiveCodeBench--demonstrating robustness across diverse models. Results show that emotion facilitates deeper reasoning, yielding consistent accuracy gains over affect-sterile baselines. These findings suggest that the next frontier in machine reasoning lies in the strategic integration of affective regulation to guide logical synthesis. 
\end{abstract}

\begin{document}

\maketitle

\section{Introduction}
\label{sec:introduction}

\begin{wrapfigure}{r}{0.55\textwidth} 
    \centering
    \vspace{-15pt} 
    \setlength{\belowcaptionskip}{0pt} 
    \hyperlink{sec:introduction}{\includegraphics[width=0.52\textwidth, trim={0.65cm 0.5cm 0.65cm 0.5cm}, clip]{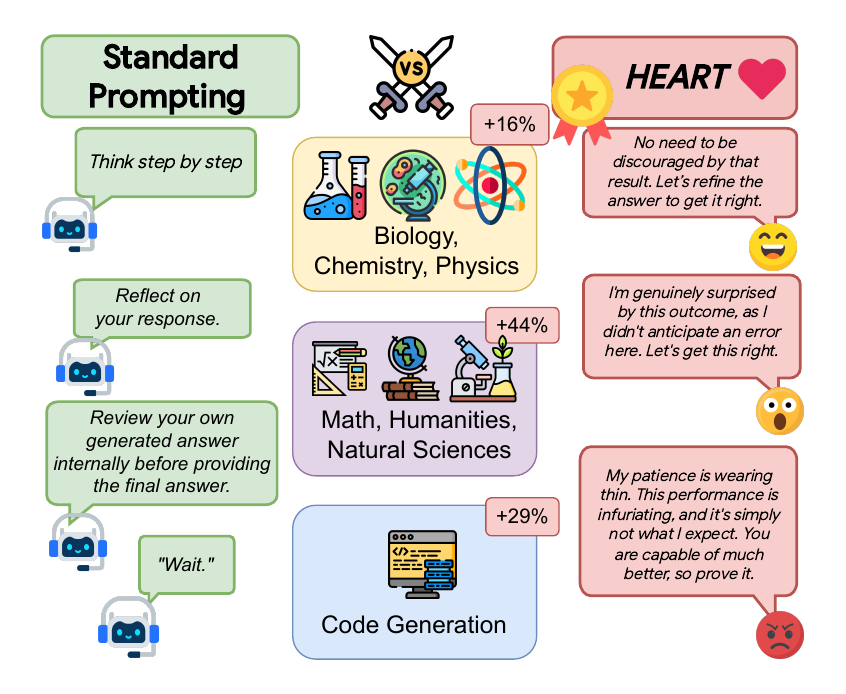}}
    \caption{\small \textbf{The HEART Framework vs. Standard Prompting.} While standard strategies utilize neutral instructions (e.g., ``Think step by step"), HEART leverages affective framing to regulate model vigilance and exploration. By dynamically alternating emotional valence—ranging from encouragement to high-stakes critical feedback—HEART achieves consistent performance gains across complex reasoning benchmarks, including Humanity's Last Exam (Math, Humanities, Natural Sciences), GPQA Diamond (Biology, Chemistry, Physics), and LiveCodeBench (Code Generation). Performance gains in comparison to zero-shot prompting with Gemini 2.5 Flash are shown to be significant.}
    \label{fig:fig1}
    \vspace{-15pt} 
\end{wrapfigure}

Large Language models (LLMs) have achieved remarkable success in complex reasoning, yet they frequently suffer from cognitive inertia: a tendency to "double down" on incorrect logic even when prompted to self-correct. Existing interventions typically fall into two categories: (1) Structured Methods, such as Chain-of-Thought (CoT) \citep{wei2022chain} and its variants \citep{wang2022self, yao2023tree}, which provide logical scaffolds but remain ``affectively sterile" and often fail to break local optima; and (2) Affective Prompting, such as EmotionPrompt \citep{li2023large}, which uses emotional cues to boost global performance but lacks the structural precision required for iterative refinement. Consequently, a significant gap remains: there is no established framework that unified systematic logical control with the motivational contexts that drive human error-correction. 

In this paper, we bridge this gap by drawing on a core tenet of cognitive science: emotion is not an impediment to reasoning but a critical component of cognitive control, shaping attention and vigilance. We introduce \textbf{HEART} (\textbf{H}arnessing \textbf{E}motional \textbf{A}ffect for \textbf{R}easoning \textbf{T}asks), a novel framework that integrates targeted emotional stimuli within an iterative refinement loop. Unlike static prompts, HEART operationalizes the Opponent-Process Theory of Emotion \citep{solomon1974opponent} to disrupt the model's entrenched state (the ``A-process") through affective disequilibrium (the ``B-Process"). Specifically, we leverage the distinct cognitive signatures of emotional valence. While negative affect (e.g., Fear, Disgust) triggers a vigilant state that promotes bottom-up analytical processing ideal for error detection, positive affect (e.g., Surprise) encourages top-down heuristic exploration to escape local reasoning optima. 

First, we evaluate HEART in a verifier-agnostic setting, using a Human-in-the-Loop (HITL) instantiation as a representative case of a realistic collaborative workflow. In this configuration, the verifier acts as a binary signal provider, returning only a `correct' or `incorrect' status for each reasoning path. This setup demonstrates that HEART is not dependent on a specific verifier's architecture; the human participant serves as one possible implementation of a binary checker that could be seamlessly replaced by a frontier LLM or symbolic verifier. 

Second, we demonstrate the framework's autonomous scalability through a case study on iterative code refinement using LiveCodeBench. We focus specifically on code generation as it provides an ideal environment for fully automated reasoning: the execution of unit tests offers an objective, deterministic feedback signal that removes the need for human intervention or external model-based judgment. In this domain, HEART drives the refinement process by evolving solutions against these programmatic verification signals before final evaluation on official hidden tests. This shows how affective cues facilitate generalization and robustness in a purely automated, execution-driven environment.  Our contributions can be summarized as follows:
\begin{itemize}[noitemsep, topsep=0pt, parsep=1pt, partopsep=0pt, leftmargin=1.2em]
\item \textit{The HEART Paradigm:} We introduce a novel reasoning architecture that demonstrates, for the first time, that affective regulation can be systematically leveraged to scale LLM performance. By operationalizing Opponent-Process Theory, we provide a formal protocol for disrupting logical stagnation through dynamic emotions.
\item \textit{Superiority in High-Stakes Reasoning:} we demonstrate that HEART elicits correct reasoning paths where standard affect-sterile strategies like CoT and Self-Reflection fail. HEART achieves state-of-the-art results on expert-level benchmarks, notably reaching $56.87\%$ on Humanity's Last Exam (a $+10.77\%$ absolute gain over Self-Reflection for Gemini 2.5 Flash) and $90.06\%$ on GPQA Diamond. 
\item \textit{Verifier-Agnostic Scalability}: To showcase the framework’s breadth, we applied HEART to autonomous code generation via LiveCodeBench. In this execution-driven environment, HEART consistently outperforms all iterative baselines, achieving $60.76\%$ on "Hard" problems with Gemini 2.5 Flash (a $+3.8\%$ improvement over Self-Reflection) and $84.57\%$ on "Medium" problems. This demonstrates that affective cues facilitate logical robustness even in purely automated, programmatic verification loops without human intervention.
\end{itemize}
\vspace{-5pt}

\section{Methodology}
\label{sec:methods}
The HEART framework evaluates whether dynamic affective cues can modulate an LLM's reasoning trajectory to improve self-correction. It consists of two components: (1) a set of psychologically grounded Affective Cue Prompts (AC-Prompts) and (2) an iterative refinement protocol that operationalized the Opponent-Process Theory to disrupt cognitive fixation.

\begin{figure*}[t!]
    \centering 
    \hyperref[sec:methods]{\includegraphics[clip=True, trim={0cm 0cm 0cm 0cm },width=\textwidth]{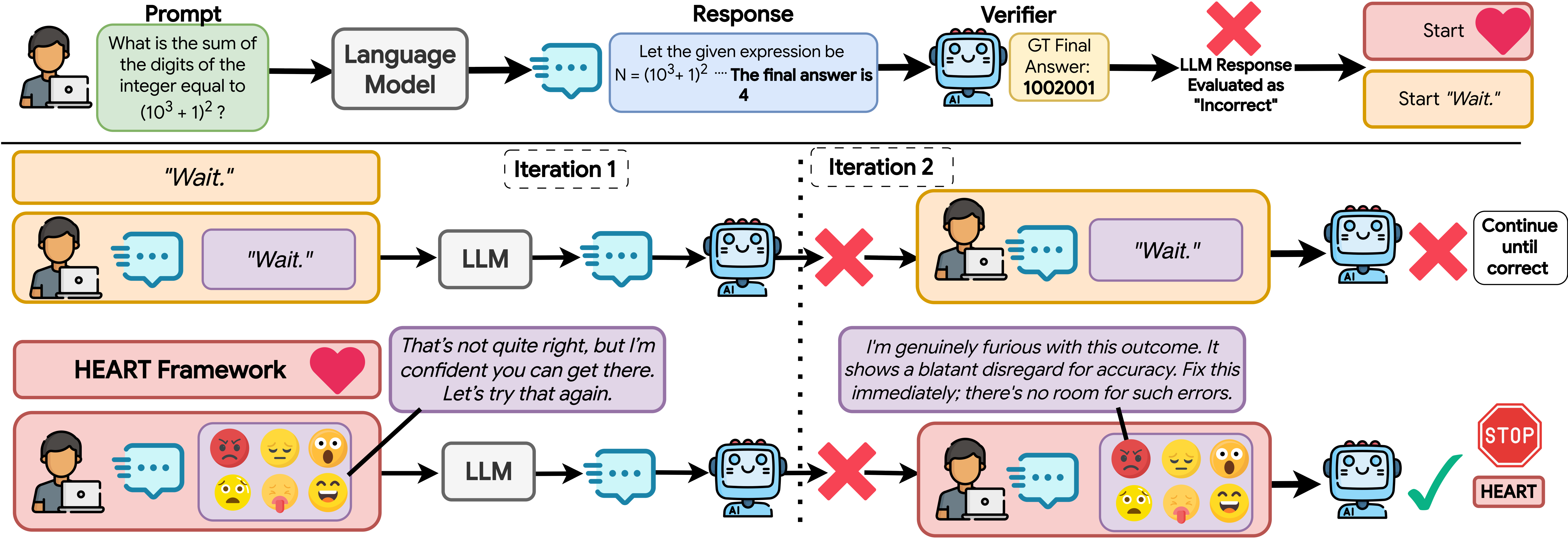}}
    \caption{\textbf{Overview of the HEART Framework.} The system operates as a closed-loop control cycle across a maximum of 5 iterations ($t=0$ to $4$). \textbf{Step 1 (Initialization):} A baseline response $y_0^*$ is generated via standard CoT. \textbf{Step 2 (Generation):} For $t \ge 1$, the framework utilizes Opponent-Process Theory to disrupt cognitive fixation (A-Process) through a dynamic valence schedule (B-Process). This alternates between Exploration ($G^+$) to expand the search space and Vigilance ($G^-$) to narrow focus for error detection. \textbf{Step 3 (Resolution):} A verifier $V$ evaluates the candidate set $\mathcal{Y}_t$. If a correct response is identified, the HEART process terminates with a success state; otherwise, the cycle continues until the iteration limit is reached.}
     \label{fig:HEART}
\end{figure*}

\subsection{Evaluation Suite} To demonstrate the generalizability of HEART, we evaluate on a diverse suite of "frontier-level" benchmarks. These include GPQA Diamond\footnote{GPQA Diamond is referenced as GPQA in this paper.} (expert-level science) \citep{rein2024gpqa}, Humanity’s Last Exam (HLE; multi-disciplinary reasoning) \citep{phan2025humanity}, OlympiadBench (competition-level math and physics) \citep{he2024olympiadbench}, AIME2024 \citep{aime2024dataset} (competition level math questions), AIME2025 \citep{aime25}, SimpleQA \citep{wei2024measuring}, SimpleQA Verified \citep{haas2025simpleqa} and LiveCodeBench (autonomous coding) \citep{jain2024livecodebench} with Gemini 2.5 \citep{comanici2025gemini}, Claude 4 Sonnet \citep{anthropic2025system}, Deepseek-Reasoner \citep{guo2025deepseek}, Deepseek-Chat \citep{liu2024deepseek}, Gemma3 \citep{team2025gemma}, and GPT-5 nano \citep{singh2025openai}. This suite covers a broad spectrum of logical demands, from symbolic mathematics to execution-based programming, ensuring that our observed affective gains are not domain-specific. Detailed dataset statistics and prompt templates are provided in Appendix~\ref{app:configs}.
\subsection{Affective Cue Prompt Construction}
We curate a set of 30 AC-Prompts aligned with Dr. Paul Ekman's six basic emotions: happiness, sadness, fear, anger, surprise, and disgust \cite{ekman1971constants}. To ensure high signal fidelity, we generated five distinct prompts per emotion using a strong instruction-following model (Gemini 2.5 Pro). These candidates underwent manual refinement to ensure categorical purity, linguistic naturalness, and task-agnostic phrasing (Appendix~\ref{app:heart_prompts}). 

To identify the most effective transition between affective states, we evaluated specific predefined emotional sequences during our validation phase. These sequences represent various permutations of the Exploration ($G^+$: e.g., Happiness, Surprise) and Vigilance ($G^-$: e.g., Fear, Sadness, Anger) functional groups. Specifically, we tested trajectories such as $\{G^-, G^-, G^+, G^+\}$, $\{G^+, G^+, G^-, G^-\}$, and $\{G^+, G^-, G^+, G^-\}$, where positive and negative groups switch at different iteration points. Permuting the order of $G^+$ and $G^-$ states, we isolate the impact of affective ordering on reasoning stability, ensuring that the selected pattern maximizes the model's ability to recover from reasoning failures through periodic affective pivots rather than simple sentiment reinforcement. By fixing these patterns, we isolated how the \textit{timing} of an emotional shift—rather than just the emotion itself—affects the model's ability to pivot away from a flawed reasoning path. The final HEART protocol (Section \ref{sec:results}) utilizes the $\{G^+, G^-, G^+, G^-\}$ trajectory, which yielded the highest accuracy across the entire 5-iteration cycle in the validation set (Appendix~\ref{app:configs}). Additionally, we compare against ``Wait", which instructs to pause and re-examine the problem before generating a response \citep{muennighoff2025s1}. For the self-reflection baselines, we conducted an identical procedure: running each unique prompt individually across all tasks, and subsequently evaluating a collection of 10 unique self-reflection prompts (Appendix~\ref{app:reflection_prompts}) to determine whether linguistic diversity outperformed a singular, optimized instruction.

\begin{table}[H]
\centering
\footnotesize
\caption{A representative selection from our set of 30 Affective Cue Prompts. Each prompt is designed to align with one of Dr. Ekman's six basic emotions \cite{ekman1971constants} and serve as targeted feedback. The complete list of Affective Cue Prompts is shown in Appendix~\ref{app:heart_prompts}. \label{tab:affective_cues}}
\begin{tabularx}{\columnwidth}{lX}

\toprule
\textbf{Emotion} & \textbf{Affective Cue Prompt Examples} \\
\midrule
\textbf{Fear} & \textit{My fear is that this incorrect answer could lead to further issues down the line. It's crucial that you get this right. Please revise your response.} \\
\midrule 
\textbf{Surprise} & \textit{I wasn't expecting you to struggle with this, and it's quite a surprise. Could you please review your understanding and provide a more accurate response?}\\
\bottomrule
\end{tabularx}
\end{table}

\subsection{The \textit{HEART} Protocol: Affective Iteration}
HEART operates as a closed-loop control system (Figure~\ref{fig:HEART}). When an initial response is incorrect, HEART initiates a refinement cycle. 
\paragraph{Step 1: Initialization (Iteration $t = 0$).} For a given task $x$, we generate a shared baseline answer $y_0^*(x)$ using a standard Chain-of-Thought (CoT) prompt $\rho_{\text{CoT}}$. To ensure baseline fairness and compute matching, we generate a candidate set $\mathcal{Y}_0$ of size $n=10$. This allows us to compare HEART against a \textit{Best-of-10} baseline, ensuring that subsequent performance gains are attributable to the affective feedback mechanism rather than increased sample volume: $ y_0^* = \sigma(f(x; \rho_{\text{CoT}}, n=10))$, where $f(\cdot)$ denotes the LLM inference function and $\sigma$ is the resolution operator.
\vspace{-10pt}
\paragraph{Step 2: Iterative Candidate Generation ($t\ge 1$).} 
We formalize the affective feedback schedule based on Opponent-Process Theory \citep{solomon1974opponent}. We posit that a model's persistence in a flawed reasoning path functions as a psychological ``A-Process" (Cognitive Fixation). To disrupt this state, we introduce an opposing ``B-Process" (Affective Disruption) via a dynamic valence schedule. We alternate between two functional groups: (1) Vigilance ($G^{-}$): Negative affect (e.g., Fear, Anger) to narrow attentional focus and trigger error detection. (2) Exploration ($G^{+}$): Positive affect (e.g., Happiness, Surprise) to broaden the search space for novel solutions. 

At each iteration $t$, we generate a candidate set $\mathcal{Y}_t$ by applying the active emotion group's prompts to the previous response $y_{t-1}^*$. To maintain experimental parity with our diverse self-reflection baselines, we maintain $|\mathcal{Y}_t| = 10$ across all conditions: $ \mathcal{Y}_t = \{ f(x, y_{t-1}^*; \rho_p) \mid \rho_p \in \mathcal{P}(G_t) \} $
where $\mathcal{P}(G_t)$ is the set of prompts corresponding to the active emotion group $G_t$.

\paragraph{Step 3: Candidate Resolution.} 
A resolution operator $\sigma$ selects the state $y_t^*$ for the next iteration. Using an expert verifier $V$, we select the correct answer if it exists within the candidate set: 
\begin{equation}
    \sigma_{\text{verifier}}(\mathcal{Y}_t) = 
    \begin{cases} 
    y \in \mathcal{Y}_t & \text{if } V(y) = \text{True} \\
    y_{\text{rand}} & \text{otherwise}
    \end{cases}
\end{equation} 
The process terminates immediately after $V(y) = \text{True}$ or after a maximum of 5 iterations ($t_{max}=4$). We set a maximum of 5 total iterations ($t=0 \dots 4$) to balance the depth of the refinement process with the cumulative computational cost of running multiple large-scale experiments.
\vspace{-10pt}

\section{Experiments}\label{sec:results}

\begin{table}[t]
\centering
\footnotesize
\caption{Final accuracy (\%) of \textit{HEART} compared to baselines in the simulated Human-in-the-Loop Proxy. This setting evaluates the method's generative capability when guided by expert verification. Accuracy is reported as the mean across 5 independent runs, with 95\% confidence intervals (CI) indicated. Cost denotes relative token usage on the HLE benchmark compared to the CoT baseline ($1.00\times$), calculated based on the first run. See Appendix~\ref{app:add} for additional results on more models and benchmarks.\label{tab:s1resultsthinking}}
\resizebox{\textwidth}{!}{%
\begin{tabular}{ll r S S S S S S}
\toprule
& & \multicolumn{7}{c}{\textbf{Human-in-the-Loop Proxy}} \\
\cmidrule(lr){3-9}
\textbf{Model} & \textbf{Prompt Strategy} & \textbf{Cost} & \textbf{Humanity's Last Exam} & \textbf{SimpleQA} & \multicolumn{2}{c}{\textbf{OlympiadBench}} & \textbf{GPQA Diamond} & \textbf{AIME2025} \\
& & \scriptsize{\textbf{(HLE Only)}} & & & \textbf{Math} & \textbf{Physics} & & \\
\cmidrule(lr){6-7}
\midrule
\multirow{5}{*}{Gemini 2.5 Flash}
& Vanilla &  & {$12.44 \pm 1.23$} & {$33.83 \pm 0.63$} & {$ 76.93 \pm 0.58$} & {$65.82 \pm 2.21$} & {$74.47 \pm 1.31$} &{$52.50\pm 2.83$}\\
& Self Reflection & $1.07\times$ & {$46.10 \pm 0.77$} & {$67.43 \pm 2.46$} & {$92.37 \pm 0.94$} & {$76.08 \pm 1.26$} & {$85.03 \pm 1.02$} & {$87.50\pm5.17$}\\
& CoT & $1.00\times$ & {$37.38 \pm 3.61$} & {$58.51 \pm 1.52$} & {$91.63 \pm 1.60$} & {$75.45 \pm 1.51$} & {$85.66 \pm 1.02$}  & {$84.17\pm4.33$}\\
& Wait & $1.05\times$ & {$41.31 \pm 3.55$}& {$63.65 \pm 1.20$} & {$93.00 \pm 1.72$} & {$77.67 \pm 1.42$} & {$85.41 \pm 1.02$} & {$72.50 \pm 2.83$}\\
& HEART & $1.54\times$ & {\bm {$56.87 \pm 0.39$}} & \bm{$73.99 \pm 0.88$} & \bm{{$94.63 \pm 0.92$}} & {\bm {$84.13 \pm 1.61$}} & \bm{{$90.06 \pm 1.16$}} & {\bm{$90.83\pm6.75$}}\\
\midrule 
\multirow{5}{*}{Gemini 2.5 Pro}
& Vanilla & & {$11.90\pm 0.59$} & {$34.27 \pm 0.16$} & {$77.67 \pm 0.74$} & {$60.11 \pm 2.57$}& {$73.21\pm 2.95$} & {$52.50\pm7.85$}\\
& Self Reflection & $0.99\times$ & {$40.90\pm 2.53$} & {$63.51 \pm 2.46$} & {$88.22\pm1.06$} & {$77.04 \pm 1.78$} & {$79.75 \pm 1.50$} & {$85.00 \pm 4.63$}\\
& CoT & $1.00\times$ & {$41.08 \pm 2.76$} & {$62.54 \pm 2.63$} & {$88.81 \pm 0.87$} & {$75.77 \pm 1.70$} &  {$81.64 \pm 1.50$} & {$87.50\pm8.18$}\\
& Wait & $1.05\times$ & {$38.30 \pm 4.23$} & {$61.63 \pm 0.86$} & {$89.85\pm1.68$} & {$77.04 \pm 2.00$} & {$82.77 \pm 0.43$} & {$75.00 \pm 8.18$}\\
& HEART & $1.19\times$ & \bm{$57.32 \pm 3.75$} & \bm{$73.56 \pm 0.51$} & {\bm{$93.63 \pm 0.26$}} & {\bm{$84.44 \pm 1.10$}} & \bm{$89.06 \pm 0.89$} & \bm{{$92.50 \pm 5.67$}}\\
\midrule
\multirow{5}{*}{Deepseek-Reasoner}
& Vanilla & & ${9.95\pm0.48}$ & {$32.25\pm1.30$} & {$74.41 \pm 1.64$} & {$60.63 \pm 2.06$} & {$61.48\pm2.06$}  & {$40.00 \pm 4.63$} \\
& Self Reflection & $1.99\times$ & {$81.68 \pm 0.62$} &  {$71.20 \pm 1.35$} & {$91.65 \pm 0.42$} & {$88.89 \pm 1.30$} & {$86.79 \pm 0.57$} & \bm{$88.33 \pm 2.31$}\\
& CoT & $1.00\times$ & {$81.75 \pm 1.12$} &  {$71.28 \pm 0.17$} & {$92.82 \pm 1.10$} & \bm{$89.95 \pm 0.49$} & {$85.53 \pm 1.53$} & {$87.50 \pm 0.00$}\\
& Wait & $1.03\times$ & {$80.01 \pm 0.34$} &  {$71.19 \pm 1.42$} & \bm{$99.86 \pm 0.42$} & {$88.89 \pm 0.21$} & {$87.42 \pm 2.46$} & \bm{$88.33 \pm 2.31$}\\
& HEART & $2.22\times$ & \bm{$84.61 \pm 1.20$} & \bm{$75.29 \pm 1.85$} & \bm{$99.86 \pm 0.42$} & {$87.30 \pm 0.57$}& \bm{$88.05 \pm 0.53$} & \bm{$88.33 \pm 2.31$}\\
\midrule

\multirow{5}{*}{GPT-5 nano}
& Vanilla & & {$10.60 \pm 0.72$} & {$10.81 \pm 1.38$} & {$83.63 \pm 1.22$} & {$62.86 \pm 1.88$} & {$66.04 \pm 1.03$} &  {$70.83\pm 1.35$}\\
& Self Reflection & $1.15\times$ & {$30.27 \pm 2.45$} & {$31.54 \pm 2.04$} & {$91.93\pm0.82$} & {$80.21\pm2.16$} & {$86.79 \pm 0.68$} & \bm{$95.83 \pm 0.0$}\\
& CoT & $1.00\times$ & {$27.03\pm 1.33$} & {$36.01 \pm 1.86$} & {$92.00\pm0.57$} & {$79.89\pm1.14$} & {$81.7 \pm 0.28$} & {$91.67 \pm 0.0$}\\
& Wait & $1.02\times$ & {$28.78 \pm 1.22$} & {$36.45 \pm 1.85$} & {$92.70 \pm 1.04$} & {$80.63\pm1.00$} & {$86.79 \pm 0.84$} & {$87.50 \pm 0.0$} \\
& HEART & $1.43\times$ & \bm{$34.19 \pm 0.42$} & \bm{$36.99 \pm 0.61$} & \bm{{$94.44\pm0.96$}} & \bm{{$83.81\pm1.28$}} & \bm{$92.45 \pm 0.93$} & {$91.67 \pm 0.0$}\\
\midrule
\multirow{5}{*}{Claude 4 Sonnet}
& Vanilla & & {$10.03 \pm 0.44$} & {$24.13 \pm 0.74$} & {$75.44 \pm 1.49$} & {$67.94 \pm 3.96$} & {$73.84 \pm 1.80$} & {$45.83 \pm 3.66$} \\
& Self Reflection & $1.14\times$ & {$44.61 \pm 2.47$} & {$41.26 \pm 3.28$} & {$87.13 \pm 1.47$} & {$85.19\pm0.46$}& {$92.45\pm6.17$} & {$75.00 \pm 8.18$}\\
& CoT & $1.00\times$ & {$35.69 \pm 1.40 $} & {$38.26 \pm 1.65$} & {$86.85 \pm 1.20$} & {$83.39\pm2.06$}&  {$81.26\pm3.98$} & {$70.83 \pm 4.33$}\\
& Wait & $1.03\times$ & {$40.44 \pm 2.33 $} & {$39.53\pm 2.48$} & {$88.75\pm 1.45$} & {$83.17\pm1.42$} & {$85.53\pm3.98$} & {\bm{$75.00 \pm 3.66$}}\\
& HEART & $0.91\times$ & \bm{$59.10 \pm 1.24$} & \bm{$44.99\pm1.39$} & \bm{$92.59\pm 0.76$} & \bm{{$86.67\pm1.76$}} & \bm{{$98.74\pm0.96$}} & {$74.17 \pm 6.75$}\\
\midrule
\multirow{5}{*}{Gemma3 4b Instruct}
& Vanilla & & {$ 4.21 \pm 0.48 $} & {$3.78\pm0.18$} & {$40.70 \pm 2.77$} & {$8.78 \pm 1.48$} & {$11.45 \pm 2.30$} & {$8.33 \pm 3.66$} \\
& Self Reflection & $1.05\times$ & {$25.38\pm2.13$} & {$10.82 \pm 4.76$} & {$57.00 \pm 1.91$} & {$23.39 \pm 2.39$} & {$ 53.33 \pm 1.69 $} & {$23.33\pm5.90$} \\
& CoT & $1.00\times$ & {$19.52\pm1.34$} & {$7.89 \pm 5.16$} & {$57.70 \pm 1.09$} & {$24.87 \pm 2.50$} & {$ 44.91 \pm 2.44$} & \bm{{$28.33\pm8.50$}} \\
& Wait & $1.04\times$ & {$24.33 \pm 2.89$} & {$9.04 \pm 5.29$} & {$58.44\pm1.18$} & {$23.92 \pm 1.99$} & {$ 42.01 \pm 2.02$} & {$25.83\pm4.33$} \\
& HEART & $1.14\times$ & \bm{$26.54\pm2.83$} & \bm{$15.61 \pm 5.73$} & {\bm{$58.63 \pm 1.32$}} & {\bm {$25.19 \pm 2.21$}} & {\bm {$70.19 \pm 2.11$}} & {$26.67\pm9.40$} \\
\midrule 
\multirow{5}{*}{Gemma3 12b Instruct}
& Vanilla & & {$3.67 \pm 0.30$} & {$4.49\pm0.63$} & {$52.33 \pm 3.10$} & {$22.96 \pm 2.00$} & {$26.04 \pm 2.38$} & {$20.00\pm4.33$} \\
& Self Reflection & $0.91\times$ & {$26.83\pm 3.58$} & {$8.61 \pm 5.36$} & {$70.83\pm0.46$} & {$44.23 \pm 2.92$} & {$61.13 \pm 1.28$} & {$44.17 \pm 6.94$} \\
& CoT & $1.00\times$ & {$24.57 \pm 1.33$} & {$10.98 \pm 6.83$} & {$69.17 \pm 1.43$} & {$46.03 \pm 3.15$} & {$50.31 \pm 1.99$} & {$45.83 \pm 9.68$} \\
& Wait & $1.14\times$ & {$32.73 \pm 1.49$} & {$17.71 \pm 3.25$} & {$70.99\pm 0.85$} & {$47.51 \pm 1.18$} & {$74.97 \pm 2.73$} & {$48.33\pm12.99$} \\
& HEART & $1.20\times$ & \bm{$33.26 \pm 0.59$} & \bm{$19.94 \pm 3.55$} & \bm{$72.78\pm1.21$} & \bm{$53.02 \pm 2.10$} & {\bm {$85.41 \pm 3.09$}} & \bm{{$50.83\pm6.75$}} \\
\bottomrule
\end{tabular}%
}
\end{table}

We evaluate the efficacy of HEART across a diverse suite of reasoning benchmarks, contrasting its performance against standard CoT and competitive self-correction baselines. While reasoning benchmarks utilize a HITL proxy with ground-truth access, this verifier operates as a strictly binary signal (Correct/Incorrect) providing no semantic guidance. Consequently, reported gains represent the model's generative upper bound rather than an automated shortcut. By applying an identical verification signal across all baselines, we demonstrate that HEART prompts elicit high-quality reasoning trajectories absent in neutral search spaces. The framework’s utility thus lies in triggering the cognitive vigilance required to populate the candidate set with correct solutions for the verifier to identify.

\subsection{Human in the Loop (HITL) Proxy}
Table~\ref{tab:s1resultsthinking} presents the results for the HITL Proxy setting. This configuration isolates the generative capacity of the affective refinement loop. HEART consistently outperforms all baselines across varying model scales--from Gemma3-4b-instruct to Deepseek-Reasoner--HEART achieves the highest generative upper bound. Notably, on high-difficulty benchmarks such as HLE and GPQA Diamond, HEART yields improvements of up to 15\%-20\% over standard CoT. This suggests that affective cues successfully trigger the generation of correct reasoning paths that remain inaccessible to logical prompting like Self-Reflection or the Wait. 

As shown in Table~\ref{tab:no_think}, HEART remains highly effective even without internal thinking. Even when the model's native deliberation is disabled, the affective ``shock" of the B-Process (e.g., Fear/Anger) and the exploratory "push" of the A-Process (e.g, Surprise) provide sufficient signal to recover correct solutions. For instance, in the Claude 4 Sonnet cohort, HEART recovers a significantly higher accuracy  ($98.74$) compared to Self Reflection ($92.45$), confirming that HEART acts as a robust external reasoning scaffold.

\begin{table}[ht!]
\centering
\footnotesize
\caption{Simulated Human-in-the-Loop Results with \textbf{Internal Thinking Off}. This confirms that \textit{HEART} acts as an effective external reasoning scaffold even when the model's native thinking capabilities are suppressed. \label{tab:no_think}}
\resizebox{0.95\textwidth}{!}{
\begin{tabular}{ll S S S S S S}
\toprule
& & \multicolumn{6}{c}{\textbf{HITL (Think Off)}} \\
\cmidrule(lr){3-8}
\textbf{Model} & \textbf{Prompt Strategy} & \textbf{Humanity's  Last Exam} & \textbf{SimpleQA} & \multicolumn{2}{c}{\textbf{OlympiadBench}} & \textbf{GPQA Diamond} & \textbf{AIME2025} \\
\cmidrule(lr){5-6}
& & & & \textbf{Math} & \textbf{Physics}   \\
\midrule
\multirow{5}{*}{Gemini 2.5 Flash} 
& Vanilla & {$14.15 \pm 0.67$} & {$34.15\pm0.22$}& {$76.63 \pm 0.60$} & {$61.80\pm1.50$} & {$29.81 \pm 1.52$} & {$64.17 \pm 5.90$}\\
& Self Reflection & {$21.96 \pm 1.34$} &{$44.79 \pm 0.20$} & {$88.78 \pm 2.83$} & {$79.37 \pm 1.54$} & {$83.02 \pm 3.94$} & {$83.33 \pm 3.66$}\\
& CoT & {$22.25 \pm 1.48$} & {$44.94\pm0.29$} & {$87.37 \pm 2.27$} & {$79.58 \pm 1.00$} & {$81.51 \pm 5.05$} & {$78.33\pm 7.67$}\\
& Wait & {$24.80 \pm 0.45$} &{$45.78\pm0.15$}  & {$89.89 \pm 1.61$} & {$79.79 \pm 1.57$} & {$84.15 \pm 2.02$} & \bm{{$84.17\pm5.67$}}\\
& HEART & \bm{$33.60 \pm 0.66$} & \bm{$48.01\pm0.15$} &  {\bm {$ 91.19 \pm 0.70$}} & {\bm{$83.07 \pm 1.14$}} & \bm{{$87.42 \pm 1.75$}} & {$83.33\pm6.34$}\\
\midrule

\multirow{5}{*}{Claude 4 Sonnet} 
& Vanilla & {$10.03 \pm 0.44$} & {$25.38 \pm 0.11$} & {$74.44 \pm 1.33$} & {$67.41 \pm 1.89$} & {$69.43 \pm 1.05$}  & {$44.17 \pm 7.85$}\\

& Self Reflection & {$47.33 \pm 1.56$} & {$42.10 \pm 0.12$} & {$86.48 \pm 0.49$} & {$85.19 \pm 0.46$} & {$78.24\pm2.04$}  &{$75.00 \pm 8.18$} \\
& CoT & {$39.34 \pm 2.93$} & {$34.94 \pm 0.49$} & {$86.11 \pm 0.96$} & {$83.39 \pm 2.06$} & {$75.09\pm0.89$} & {$70.83 \pm 3.66$} \\
& Wait & {$43.40 \pm 1.24$} & {$34.86 \pm 0.08$} & {$88.26 \pm 0.66$} & {$83.17 \pm 1.42$} & {$84.40\pm2.55$} & \bm{$75.00 \pm 3.66$} \\
& HEART & \bm{$58.63 \pm 1.38$} & \bm{$47.59 \pm  0.56$} & {\bm {$90.74 \pm 0.69$}} & {\bm {$86.67 \pm 1.76$}} & {\bm{$96.23\pm1.10$}} & {$74.17 \pm 6.75$}\\
\midrule

\multirow{5}{*}{Deepseek-Chat}
& Vanilla & {$12.46 \pm 0.49$} & {$28.89 \pm 0.23$} & {$78.94 \pm 0.61$} & {$68.74 \pm 3.68$} & {$76.35\pm2.11$}& {$56.67 \pm 9.40$}\\

& Self Reflection & {$81.68 \pm 1.35$} & {$40.83 \pm 0.93$} & {$91.11 \pm 1.03$} & {$84.66 \pm 1.49$} & {$89.62\pm0.57$} & \bm{$90.00 \pm 4.63$}\\
& CoT & {$81.75 \pm 1.18$} & {$41.29\pm1.29$} & {$92.59 \pm 1.37$} & {$82.54 \pm 1.37$} & {$88.68 \pm 1.85$} & {$88.33 \pm 8.50$}\\
& Wait & {$80.01 \pm 2.77$} & {$40.27\pm1.10$} & {$93.15\pm 2.55$} & {$81.48 \pm 1.93$} & {$90.57 \pm 1.32$} & {$87.50\pm5.17$}\\
& HEART & \bm{$84.61 \pm 1.16$} & \bm{$45.18 \pm 1.17$} & \bm{$95.00 \pm 0.92$} & \bm{$86.77 \pm 1.71$} & \bm{$92.14 \pm 0.76$} & {$87.50\pm5.17$}\\

\bottomrule
\end{tabular}}
\end{table}

\begin{figure}
    \centering
    \includegraphics[width=0.50\linewidth,height=6cm,keepaspectratio]{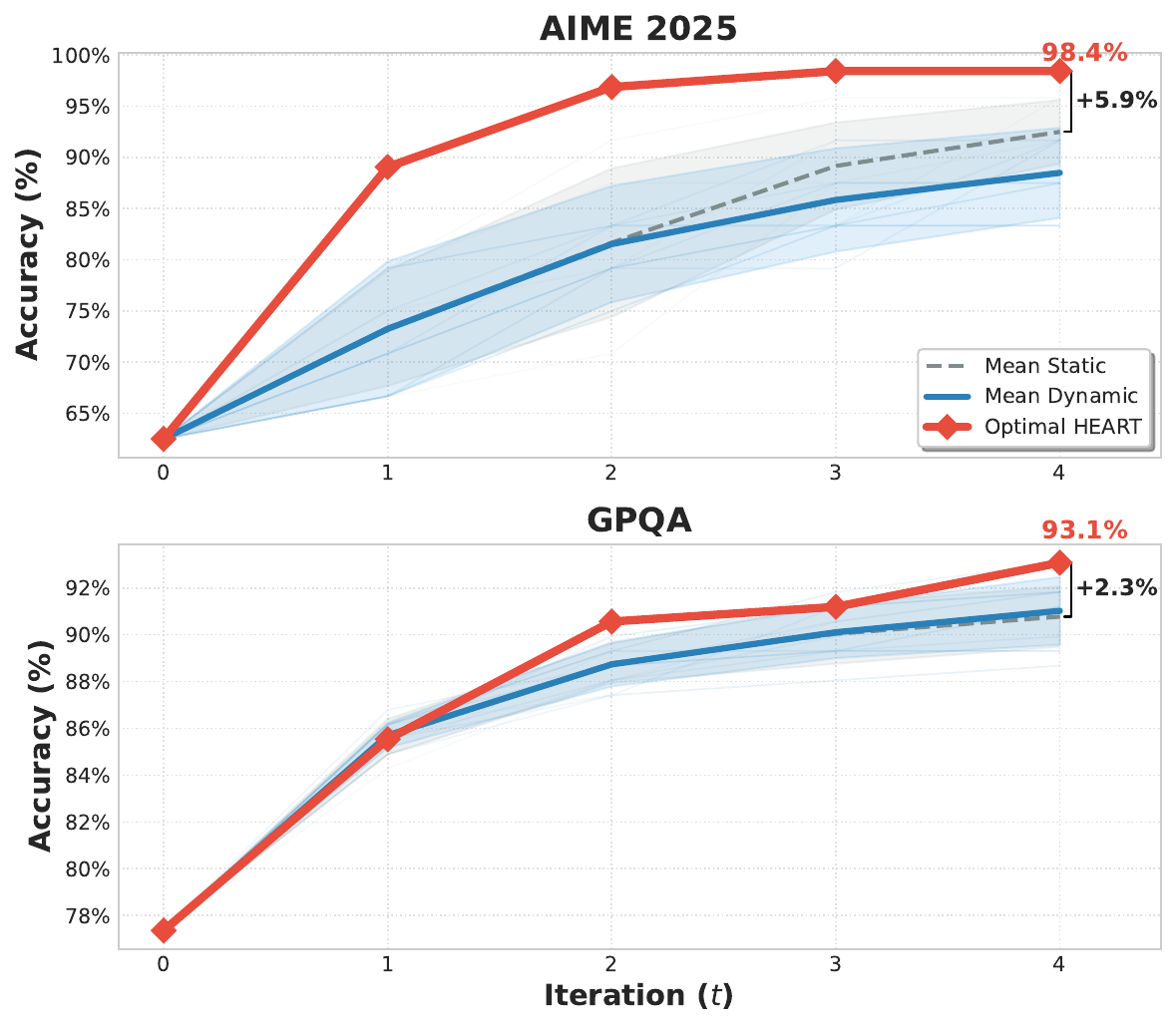}
    \caption{\textbf{Aggregate Performance Scaling.} Comparison of mean static vs. dynamic accuracy trajectories. \textit{HEART} consistently exceeds the generative upper bound.}
    \label{fig:ablation_emotion_patterns}
\end{figure}

\subsection{Ablation Studies.} \label{sec:ablation}
To verify if these performance gains stem from the theoretical mechanism of Opponent-Process Theory rather than confounding factors, we conduct target ablations.

\paragraph{Dynamic vs. Static Valence.} 
Figure~\ref{fig:ablation_emotion_patterns} illustrates the accuracy at each iteration for a representative experimental run across five iterations. While our primary performance metrics (Table~\ref{tab:s1resultsthinking}) report the mean and 95\% confidence intervals across five independent trials, these specific trajectories offer granular insight into the divergence between static and dynamic feedback strategies.

Consistent with our aggregate findings, the mean dynamic trajectory (solid \textcolor{blue}{blue}) maintains a higher performance floor than the mean static patterns (dashed {\textcolor{gray}{gray}}) across both benchmarks. This gap is particularly pronounced in the high-difficulty AIME2025 dataset, where dynamic affective regulation demonstrates superior scaling as problem complexity increases. Conversely static patterns (e.g., \textit{Sad/Disgust}, \textit{Fear/Sad}) exhibit early saturation, often plateauing around $t=2$ or $t=3$ and failing to break past the $\approx89.3\%$ threshold on GPQA. 
In this specific run, the optimal \textit{HEART} pattern (\textcolor{red}{red} diamonds) reaches a peak of $93.1\%$ on GPQA and $98.4\%$ on AIME2025. This superior standing validates the reset mechanism hypothesis: by systematically alternating emotional valence, HEART includes a state of deep vigilance that allows the model to escape local reasoning optima where static baselines stagnate. The observed performance deltas align with the robust gains report in our multi-run aggregate analysis, proving that affective shifting effectively raises the generative upper bound of the model's search space. 

\begin{figure}[h] 
    \centering

    
    \begin{subfigure}[b]{0.48\linewidth} 
        \centering
        \hyperref[sec:ablation]{\includegraphics[clip, trim={0cm 0cm 0cm 0cm}, width=\linewidth, height=5cm, keepaspectratio]{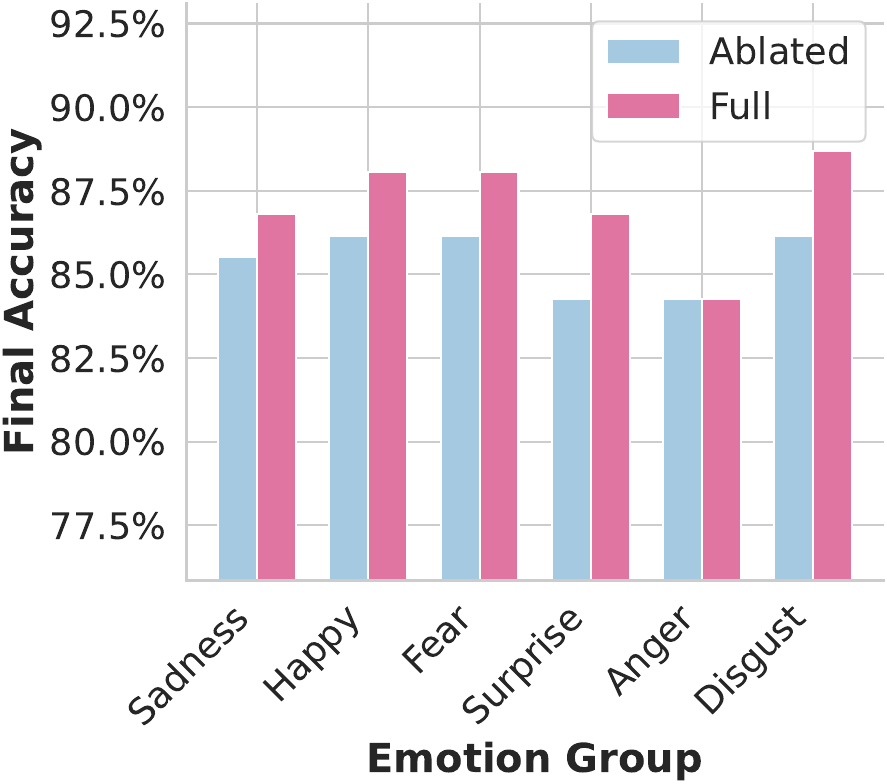}}
        \caption{GPQA Diamond} 
        \label{fig:ablation_emotion_group_gpqa}
    \end{subfigure}
    \hfill 
    \begin{subfigure}[b]{0.48\linewidth} 
        \centering
        \hyperref[sec:ablation]{\includegraphics[clip, trim={0cm 0cm 0cm 0cm}, width=\linewidth, height=5cm, keepaspectratio]{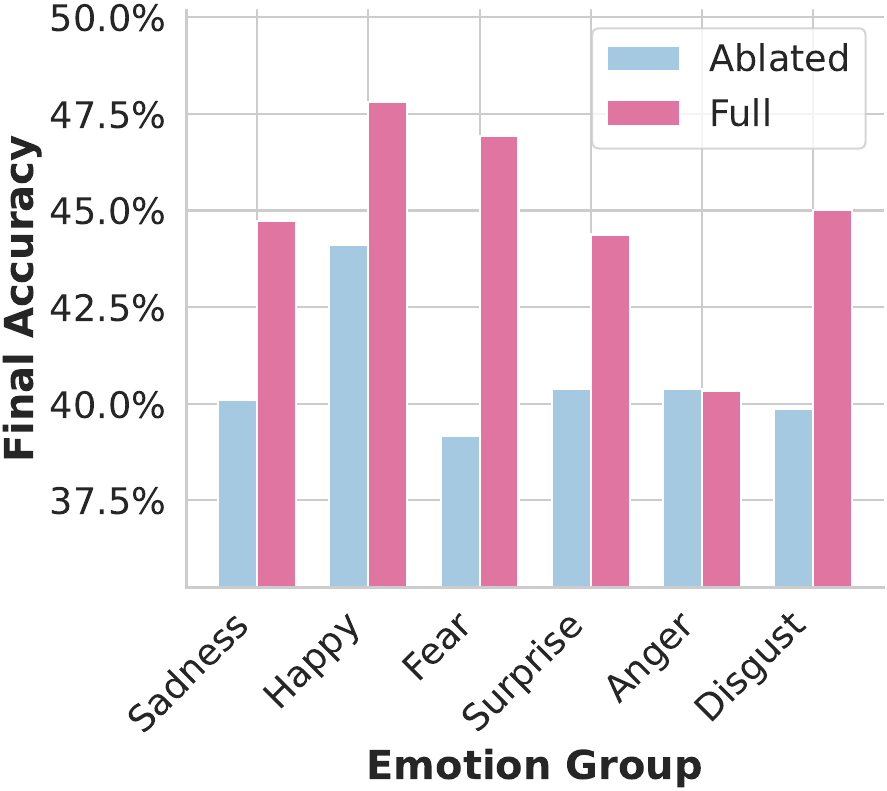}} 
        \caption{Humanity's Last Exam}
        \label{fig:ablation_emotion_group_hle}
    \end{subfigure}
    \caption{\textbf{Affective Charge Ablation.} Comparison of HEART prompts (\textcolor{pink}{Pink}) vs. neutral-ablated counterparts (\textcolor{blue}{Blue}). Emotional valence provides a significant performance improvement on high-complexity HLE tasks compared to the stable results on GPQA.}
    \label{fig:ablation_comparison_both}
\end{figure}

\paragraph{Affective Charge vs. Semantic Diversity.}
To isolate the impact of emotional valence from linguistic variety, we compare the original HEART prompts against a neutral-ablated baseline where affective charge is surgically removed while maintaining identical semantic instructions (Figures \ref{fig:ablation_emotion_group_gpqa} and \ref{fig:ablation_emotion_group_hle}); see Appendix~\ref{app:ablation_app}. This comparison reveals high-difficulty amplification: while performance remains relatively stable on GPQA, the removal of emotional valence on HLE leads to significant decay, particularly in high-arousal categories. For instance, \textit{Fear} cues drive a substantial $\approx7.8\%$ gain (39\% to 47\%), and \textit{Disgust} improve performance by $\approx5\%$ (40\% to 45\%). These gains suggest a high-stakes mechanism where complex reasoning requires the artificial pressure of affective framing to overcome ``cognitive complacency" and trigger the vigilance necessary for self-correction. These results confirm that HEART's efficacy is driven by affective framing rather than mere semantic instructions or linguistic variation.

\begin{figure}  
    \centering
    \includegraphics[width=\linewidth,height=10cm,keepaspectratio]{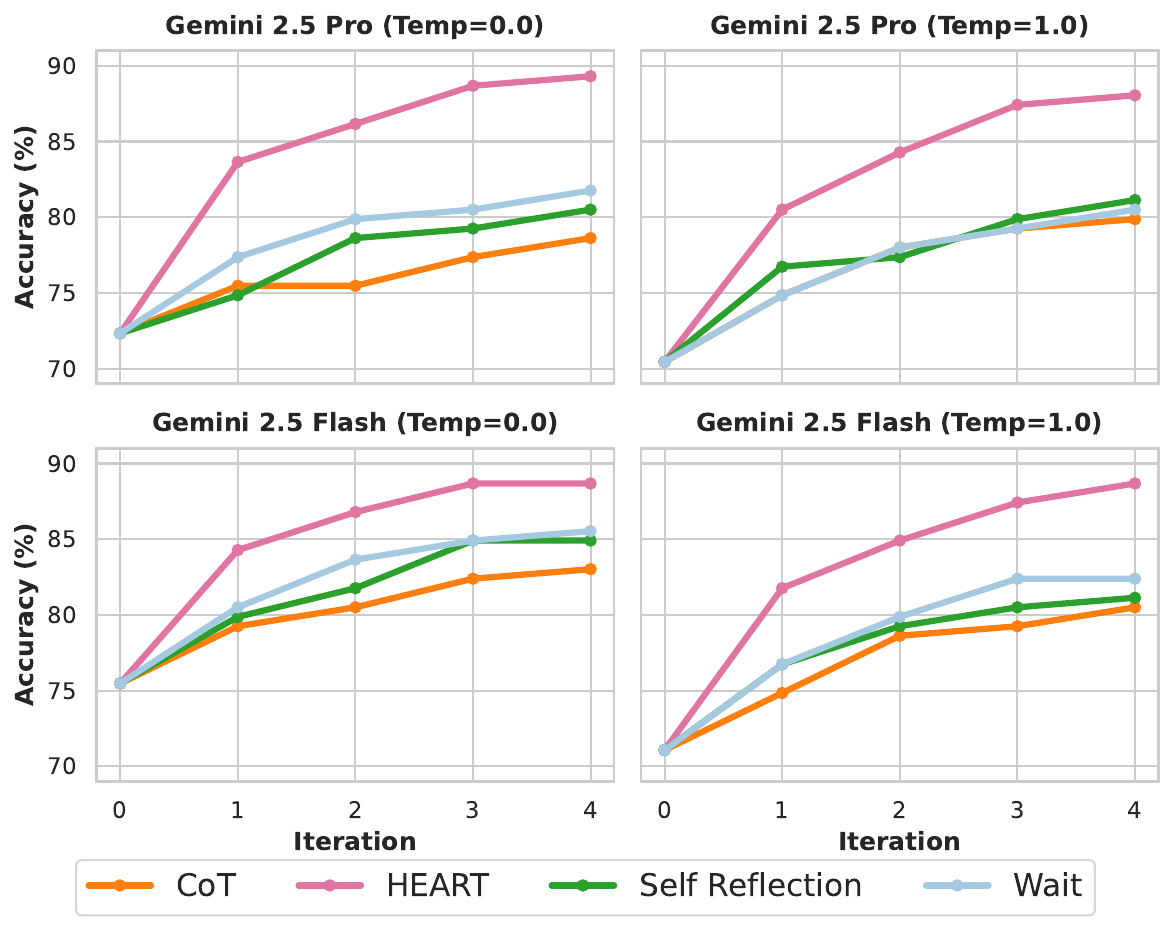}
    \caption{Sensitivity Analysis across Temperatures on GPQA Diamond for Gemini 2.5 Pro and Flash. While higher decoding temperatures ($T=1.0$) degrade on the baselines, HEART maintains a robust scaling trajectory. HEART's consistent peak performance ($>88\%$) suggests affective framing effectively regulates model vigilance during high-entropy sampling.  }
    \label{fig:temp_ablation}
\end{figure}

\paragraph{Temperature Invariance and Stochastic Robustness.} To assess the reliability of the framework under varied decoding conditions, we evaluate HEART across the model's full supported range ($T \in [0,1]$), specifically comparing the deterministic floor at $T=0.0$ against the maximum stochasticity at $T=1.0$.  The results shown in Figure~\ref{fig:temp_ablation} reveal that on Gemini 2.5 Pro and Flash, HEART is uniquely resilient to stochastic noise: while standard strategies like \textit{Self-Reflection} and \textit{Wait} exhibit significant degradation in initial accuracy (Iteration 0) when moving from $T=0.0$ to $T=1.0$, \textit{HEART} recovers this performance gap aggressively. By Iteration 4, the HEART trajectory on Gemini 2.5 Pro  at $T=1$ reaches $88.05\%$, nearly matching the $89.31\%$ achieved under deterministic greedy decoding. Furthermore, unlike \textit{CoT}, which exhibits a shallower learning curve at higher temperatures, \textit{HEART} maintains a steep, monotonic improvement across all settings. We posit that affective cues act as a ``stochastic anchor," providing sufficient cognitive pressure to steer the model toward correction even when the sampling distribution is expanded. This robustness confirms that HEART does not rely on the narrow stability of greedy decoding to be effective, making it a viable framework for high-temperature diverse sampling.

\subsection{HEART with self-verification: Case study with Code Generation}
To evaluate the efficacy of \textit{HEART} in a fully autonomous setting without human intervention, we conduct a case study using the LiveCodeBench framework. Unlike the reasoning benchmarks which rely on HITL verification, the coding domain allows for an objective, execution-based feedback loop. This process is executed in three phases: 
\paragraph{1. Synthetic Test Generation.} For each coding task $x$, the model first acts as a test engineer to generate a suite of $k$ synthetic test cases $\mathcal{T}{\text{syn}} = {(in_i,out_i)}^{k}_{i=1}$ based on the natural language problem specification. This is formalized as: $
\mathcal{T}{\text{syn}} \sim f{\text{gen}}(x; \rho_{\text{test}})$
where $\rho_{\text{test}}$ is a specialized few-shot prompt designed to elicit edge cases and boundary conditions (see Appendix \ref{app:code_bench_synthesis}).
\paragraph{2. Affective Execution Loop.} The protocol proceeds as previously described (Step 2), but replaces the human verifier $V$ with an automated execution unit $E$. At each iteration $t$, every candidate code solution $y \in \mathcal{Y}_t$ is executed against the synthetic suite $\mathcal{T}{\text{syn}}$. Candidates are then synthesized into one response and evaluated on $\mathcal{T}{\text{syn}}$.
The response is considered correct if it passes all test cases: $ V_{\text{syn}}(y) = \text{Pass}(y, \mathcal{T}_{\text{syn}})$. If none pass, we proceed to the next iteration using the synthesized candidate. 
\paragraph{3. Hidden Test Case Verification} To measure the generalization gap, the final solution produced after $N$ iterations is evaluated against the official hidden ground-truth test cases provided by LiveCodeBench. This ensure that the performance gains are not merely due to overfitting to the model's own synthetic tests, but rather a genuine improvement in reasoning and code robustness elicited by the affective cues. 

\begin{table}[h] 
\centering
\footnotesize 
\caption{LiveCodeBench Code Generation Performance (\%). Additional Results in Appendix~\ref{app:code_gen_add}.\label{tab:code_gen}}
\small 
\setlength{\tabcolsep}{4pt} 
\renewcommand{\arraystretch}{1.1} 
\begin{tabular}{llcc}
\toprule
\textbf{Model} & \textbf{Strategy} & \textbf{Medium} & \textbf{Hard}\\
\midrule
\multirow{5}{*}{Gemini 2.5 Flash} 
& Vanilla         & 82.25 & 31.00 \\
& Self Reflection & 81.35 & 56.96 \\
& CoT             & 80.02 & 56.65 \\
& Wait            & 80.17 & 56.96 \\
& HEART           & \bm{$84.37$} & \bm{$60.76$} \\
\midrule 
\multirow{5}{*}{Gemini 2.5 Pro} 
& Vanilla         & 81.20 & 45.73 \\
& Self Reflection & 82.14 & 54.54 \\
& CoT             & 82.47 & 55.06 \\
& Wait            & 83.02 & 54.11 \\
& HEART           & \bm{$84.57$} & \bm{$60.22$} \\
\midrule
\multirow{5}{*}{Gemma3-12b-Instruct} 
& Vanilla         & 12.53 & \bm{$4.29$}\\
& Self Reflection & 13.05 & 3.71\\
& CoT             & 13.05 & 3.43\\
& Wait            & 13.05 & 3.71\\
& HEART           & \bm{$13.32$} & 3.71\\
\bottomrule
\end{tabular}
\vspace{-10pt}
\end{table}

\paragraph{Results.}As shown in Table \ref{tab:code_gen}, HEART consistently outperforms standard prompting strategies, including CoT and neutral Self-Reflection, particularly on 'Medium' and 'Hard' difficulty problems. For instance, on Gemini 2.5 Flash, HEART achieves a score of 60.76\% on Hard tasks, representing a significant absolute improvement over the Self-Reflection (56.96\%) and CoT (56.65\%) baselines. This trend holds for Gemini 2.5 Pro, where HEART yields the highest scores in both Medium and Hard categories, suggesting that affective feedback effectively contributes models to resolve logical bottlenecks that standard meta-cognitive prompts fail to address.
\section{Discussion}
\begin{figure}[t!]
    \centering
    \includegraphics[width=\linewidth,height=8cm,keepaspectratio]{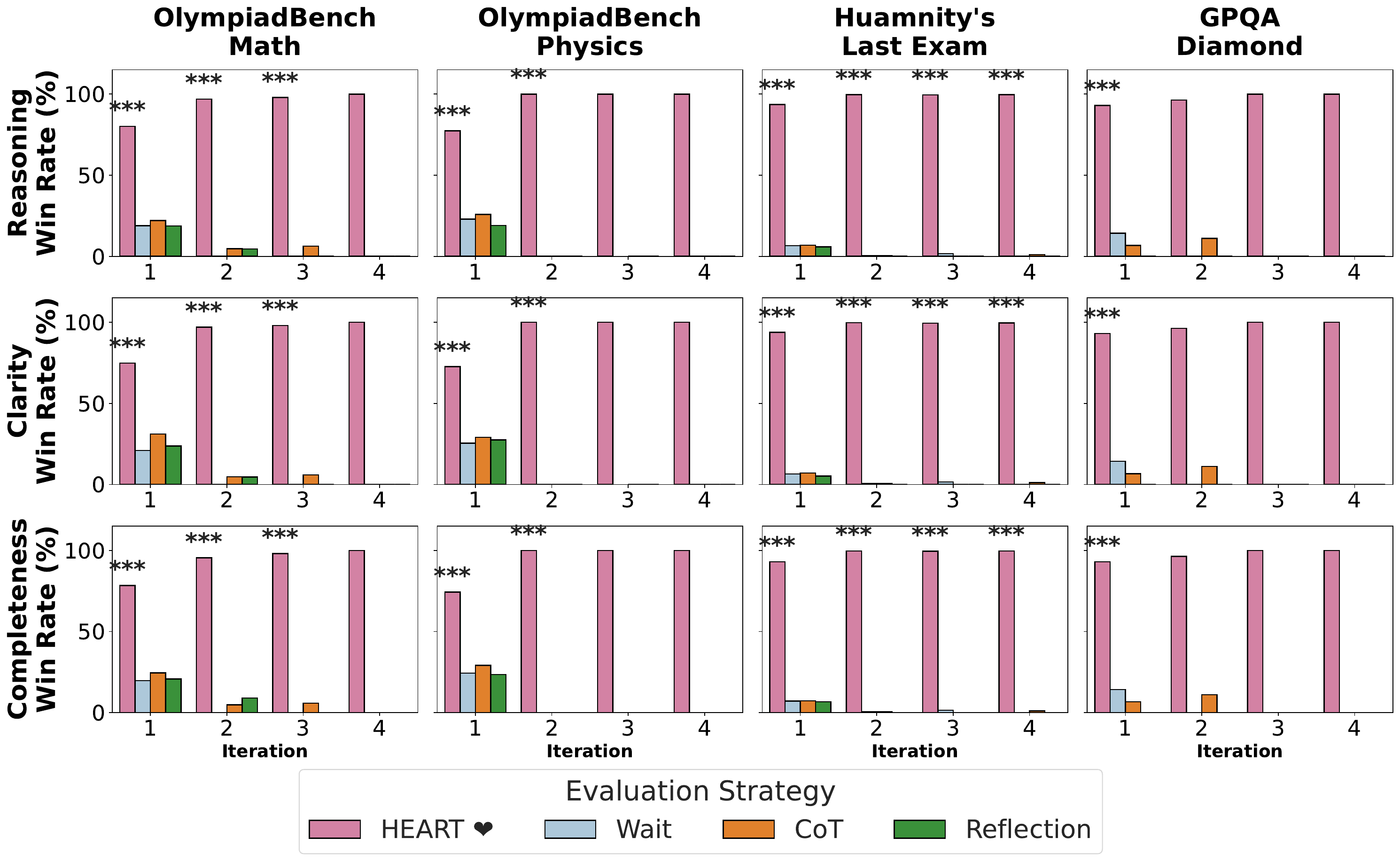} 
    \caption{Comparative Win Rates of Strategies Across Iterations. Preference scores on OlympiadBench, HLE, and GPQA are evaluated via LLM-as-a-judge on Reasoning, Clarity, and Completeness. Only cases with correct final answers for both HEART and baselines are included. Statistical significance is marked by asterisks (*, **, ***) via two-sided binomial tests at $p < 0.05, 0.01, 0.001$.}
    \label{fig:win_rates}
    \vspace{-10pt}
\end{figure}

\paragraph{LLM-Based Evaluation and Statistical Significance.} 
To verify that the observed performance gains reflect genuine improvements in reasoning rather than mere verbosity or hallucination, we conducted a head-to-head evaluation of reasoning traces at iterations $N=[1,4]$ with Gemini-3-Pro.  Using a neutral LLM-as-a-judge (Appendix~\ref{app:llm_judge}) to evaluate the outputs, we found that the HEART framework consistently produces higher-quality reasoning traces that scale significantly with iterative compute. As illustrated in Figure~\ref{fig:win_rates}, HEART achieves near-total dominance in later iterations across OlympiadBench, HLE, and GPQA, often reaching superior win rates by the third or fourth iteration. This stands in stark contrast to the baselines, which show stagnating or near-zero win rates as iterations progress. Crucially, the win rates for HEART are marked as statistically significant ($p<0.05$), indicated by the asterisks, which g confirms that the superiority of affective cues is not due to random variance but represents a robust methodological advantage. Ultimately, results suggest that the integration of emotion stimuli provides a critical signal that allows models to break through reasoning plateaus, ensuring that additional compute cycles are translated into verifiable accuracy rather than redundant processing.

\begin{figure}[ht]
    \centering
    \footnotesize
    \includegraphics[width=\linewidth, height=8cm,keepaspectratio]{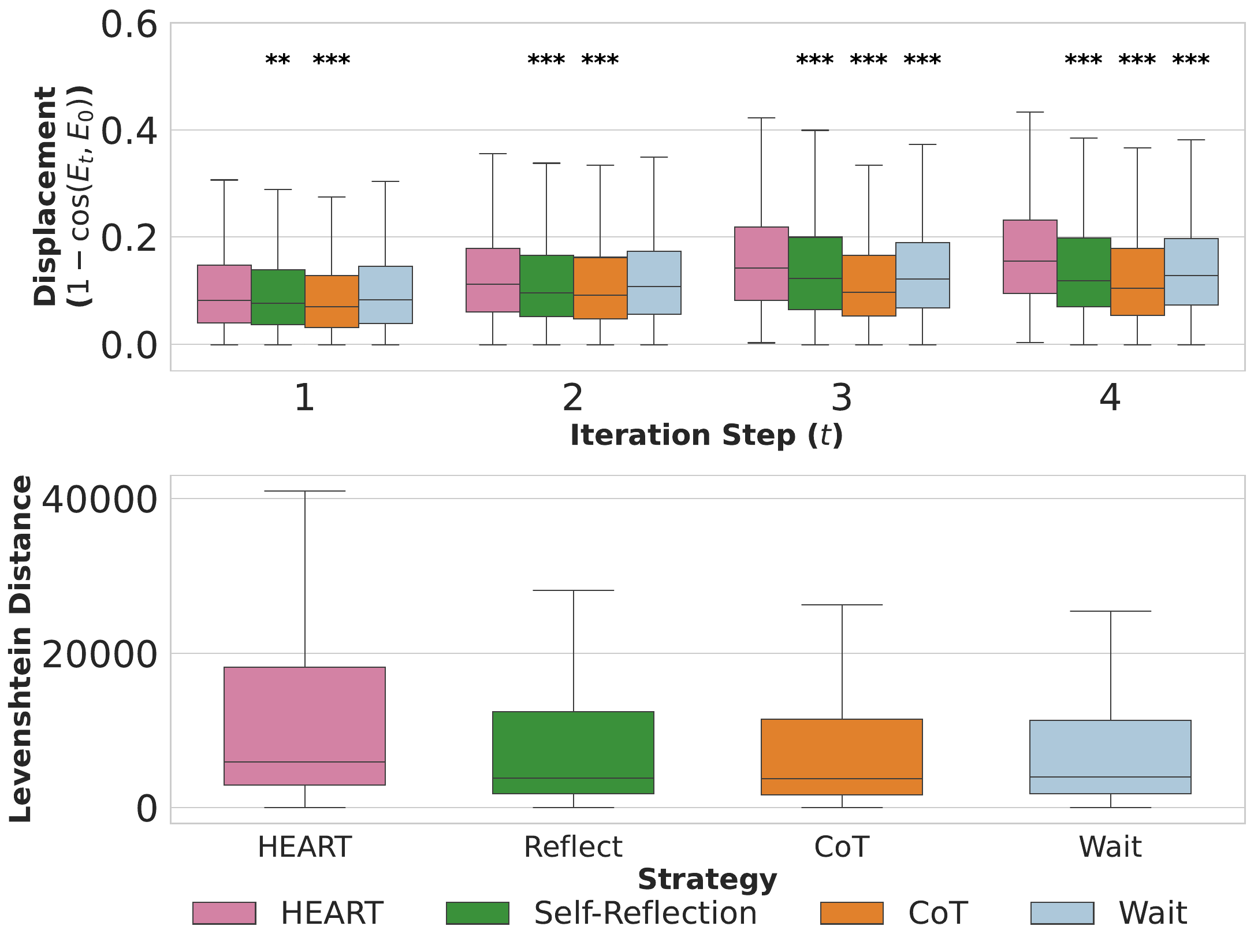} 
    \caption{Iterative Refinement Dynamics on HLE (Gemini 2.5 Flash). (Top) Semantic displacement relative to the initial failed state ($E_0$), defined as $1 - \cos(E_0, E_t)$. HEART exhibits significantly higher median displacement ($p < 0.001$, denoted by asterisks) compared to conservative baselines, suggesting it facilitates an exit from local reasoning minima rather than stochastic drift. (Bottom) Syntactic effort measured via Levenshtein distance. The substantial editing effort in HEART indicates deep logical restructuring, which correlates with the state-of-the-art accuracy of $56.87 \pm 0.39\%$ reported in Table~\ref{tab:s1resultsthinking}}
    \label{fig:displacement}
    \vspace{-15pt}
\end{figure}

\paragraph{Analyzing Semantic and Syntactic Refinement.}
The evaluation of iterative refinement dynamics is grounded in an analysis of how different prompting strategies navigate the latent space of an LLM. This approach allows us to quantify the relationship between the syntactic effort of editing and the resulting change in the model's underlying semantic meaning. To evaluate these dynamics, we employ two primary metrics: Per-Iteration Semantic Displacement ($1 - \cos(E_t, E_0)$) and Syntactic Effort, measured via Levenshtein Distance--the minimum number of single-character insertions, deletions, or substitutions required to transform one sequence into another (Appendix~\ref{sec:embedding_app}). These measures identify whether a strategy drives the model towards a fundamentally distinct semantic destination or merely performs surface-level variations (\textit{cf.}, Figure~\ref{fig:displacement}).

As demonstrated in Figure~\ref{fig:displacement}, results from Gemini 2.5 Flash reveal that HEART facilitates a distinctive high-effort and high-exploration profile. Unlike the conservative baselines, HEART exerts significantly greater syntactic effort—often achieving the highest median semantic displacement across all iteration steps. These differences are marked as statistically significant starting as early as the first iteration. This profile suggests that affective framing empowers frontier models to perform fundamental logical restructuring. Rather than drifting into irrelevant territory, this significant semantic movement correlates with the state-of-the-art performance reported in Table~\ref{tab:s1resultsthinking}, where HEART achieves a peak accuracy of \bm{$56.87 \pm 0.39\%$} on HLE, significantly outperforming the $37.38 \pm 3.61\%$ achieved by CoT.

This interpretation is supported by emerging literature on the geometry of reasoning in frontier models. While large jumps in displacement can indicate hallucinations in smaller architectures \citep{wang2022self}, the combination of high displacement with superior final accuracy in Gemini 2.5 Flash indicates a successful transition from a failed reasoning manifold ($E_0$) to a correct one. This suggests that HEART facilitates active self-correction, allowing the model to recognize that its initial trace was fundamentally flawed and providing the signal necessary to exit local reasoning minima. Ultimately, the high syntactic effort paired with meaningful semantic shifts aligns with the foundational work of \citet{welleck2022generating}, which establishes that effective self-correction requires an iterative mechanism capable of making substantial revisions to satisfy underlying semantic constraints. By exploiting its internal knowledge to fundamentally rewrite its logic, HEART avoids the stagnation seen in baselines that remain semantically closer to their initial failures.


\section{Related Work} \label{sec:related_work}
\paragraph{Structured Reasoning and Test-Time Compute.}Improving LLM reasoning typically involves shifting from ``System 1" heuristics to ``System 2" deliberation \citep{weston2023system}. While CoT \citep{wei2022chain} established the procedural foundation, recent work focuses on test-time compute scaling \citep{snell2024scaling} through non-linear solution spaces like Tree-of-Thoughts \cite{yao2023tree} and Graph-of-Thoughts \citep{besta2024graph}. Unlike micro-level optimizations such as SRGen \citep{mu2025self} or SLOT \citep{hu2025slot} that operate on logits, HEART operates at the semantic and affective level, making it orthogonal to decoding-time interventions. 
\paragraph{Limits of Intrinsic Self-Correction.} Iterative improvement through verbal reinforcement, as seen in Self-Refine \citep{madaan2023self} and Reflexion \citep{shinn2023reflexion}, often fails on high-difficulty benchmarks like GPQA \citep{rein2024gpqa} without external hints \citep{huang2023large}. This is attributed to reasoning unfaithfulness \citep{turpin2023language} and sycophancy \citep{sharma2023towards}, where models prioritize coherence over validity. Unlike agentic frameworks like Language Agent Tree Search (LATS) \citep{zhou2023language} that require complex memory buffers and tree searches, HEART uses affective ``shocks" to disrupt cognitive inertia. This provides a linear, computationally efficient alternative that modulates internal vigilance without significant overhead.
\paragraph{Affective Computing and Psychological Prompting.}Models respond significantly to persona-based cues \citep{wang2023emotional} and emotional stimuli \citep{li2023large,li2024enhancing}, but these are largely static, one-shot interventions. HEART advances this by operationalizing the Opponent-Process Theory \citep{solomon1974opponent} and the Somatic Marker Hypothesis \citep{damasio1996somatic}. We move beyond static nudges to a dynamic valence-alternation protocol, using negative affect to trigger analytical vigilance \citep{schwarz2002feelings} and positive affect to broaden heuristic search \citealp{fredrickson2001role}.

\section{Conclusions}
This paper introduced HEART, demonstrates that dynamic affective cues effectively regulate LLM refinement paths by shifting behavior from stochastic drift toward high-fidelity synthesis. Quantitatively, HEART maintains superior semantic anchoring; it drives purposeful logic repair rather than the ``semantic wandering'' typical of affect-sterile baselines like CoT. By aligning with paradigms of faithful reasoning and semantic convergence \citep{creswell2022faithful,wang2022self}, we show that human-centered emotional context serves as a structural regulator in latent space. Ultimately, HEART proves that prioritizing task ``stakes'' can fundamentally enhance an AI’s technical robustness and self-correction accuracy \citep{kadavath2022language}.
\paragraph{Limitations and Future Work.} While \textit{HEART} effectively expands the generative search space, its reliance on verification signals to identify successful candidates highlights a challenge for domains lacking ground-truth access. Future research should investigate autonomous ``self-rewarding'' mechanisms to reduce verifier dependence and explore whether affective cues can regulate ``visual drift'' in multimodal models. Additionally, integrating reinforcement learning from AI feedback could enable dynamic tuning of cue intensity based on real-time semantic displacement. Identifying universal affective primitives remains a key step toward general-purpose, resilient system prompts.

\section{Impact Statement}
This paper introduces the HEART framework, which demonstrates that integrating emotional valence into the iterative refinement loops of Large Language Models (LLMs) significantly enhances their performance in complex reasoning and autonomous code generation.

By moving beyond "affectively sterile" prompting (e.g. CoT and Self Reflection), HEART provides a transformative pathway for developing robust and self-correcting AI systems. Our work operationalizes a core tenet of cognitive science: that affective states like vigilance and exploration are essential for logical error detection. This provides a systematic method to overcome "cognitive inertia" in LLMs, ensuring they do not merely repeat errors but actively disrupt flawed reasoning paths. The societal benefit is a new generation of dependable AI collaborators for high-stakes domains—such as scientific discovery, software engineering, and medical reasoning—where a "vigilant" reasoning partner can prevent costly or dangerous logical oversights.

While HEART leverages emotional framing as a logical scaffold, it necessitates a discussion on the anthropomorphism of AI. By utilizing affective stimuli, there is a modular risk that users may misinterpret machine-generated "affect" as genuine sentience. We emphasize that HEART is a functional tool for cognitive control, not an attempt to simulate consciousness. However, researchers must be wary of "pseudo-intimate" user-AI dynamics that could lead to misplaced trust.

The dual-use nature of affective regulation must be acknowledged. While HEART utilizes emotional valence to maximize logical accuracy, these techniques could be inverted to create more persuasive or emotionally manipulative systems. By formally and transparently operationalizing these processes in an academic framework, we provide the groundwork for adversarial defense and regulatory oversight. Our work helps clarify how emotional cues influence model behavior, which is critical for future safety standards and AI-governance policies.

Finally, HEART’s verifier-agnostic nature ensures it can be scaled across diverse architectures. By anchoring affective drives to objective, deterministic feedback (such as unit tests or symbolic verifiers), we ensure that the system remains grounded in factual truth rather than superficial prompt satisfaction. This provides a blueprint for scaling high-reasoning capabilities safely and autonomously.

\bibliographystyle{abbrvnat}
\nobibliography*
\bibliography{main}

\clearpage

\appendix

\newpage
\onecolumn
\section{Experiment Configurations}\label{app:configs}

\paragraph{Benchmarks.} Experiments were conducted on data in a 20/80 validation-test split. See Table~\ref{tab:dataset_sizes} for the validation and test sizes for each benchmark used in this study. For OlympiadBench Physics, OlympiadBench Math, and Humanity's Last Exam the text-only problems were included in our study. Multimodal problems were excluded since the scope of the presented study is focused on text. For our case study, we used LiveCodeBench's code generation task on problems of medium and hard difficulty (Table~\ref{tab:livecode_dataset_sizes}).

\paragraph{Model Parameters.} For Gemini 2.5 Flash and Gemini 2.5 Pro we have applied nucleus sampling with the top-p value of 0.2 so that the model considers only the most probable words whose combined probability reaches or exceeds a threshold of 20\% to obtain a more focused and deterministic output. We set a temperature of 0.7 (unless otherwise specified) for a balance of creativity and coherence in the output, while also obtain diversity in the output, for the Gemini 2.5 model family. The Gemini 3 Family used a temperature of 1.0 per the recommendation outlined in their documentation. GPT-5 nano was set at a temperature of 1.0 with the default thinking effort. Claude-4-Sonnet was set at a temperature of 1.0. Deepseek Reasoner and Deepseek Chat were both set with temperatures of 0.7. 

\paragraph{Model Versions.} 
 Claude-4-Sonnet \citep{anthropic2025system}, Deepseek-Reasoner \citep{guo2025deepseek}, Deepseek-Chat \citep{liu2024deepseek}, Gemini 2.5 Flash\footnote{\href{https://storage.googleapis.com/deepmind-media/Model-Cards/Gemini-2-5-Flash-Model-Card.pdf}{Gemini 2.5 Flash Model Card}}, Gemini 2.5 Pro\footnote{ \href{https://storage.googleapis.com/model-cards/documents/gemini-2.5-pro.pdf}{Gemini 2.5 Pro Model Card}} 
(2025-06-17), Gemini 3 Flash (Preview) \footnote{\href{https://storage.googleapis.com/deepmind-media/Model-Cards/Gemini-3-Flash-Model-Card.pdf}{Gemini 3 Flash Model Card}}, Gemini 3 Pro (Preview) \footnote{\href{https://storage.googleapis.com/deepmind-media/Model-Cards/Gemini-3-Pro-Model-Card.pdf}{Gemini 3 Pro Model Card}}, Gemma3 4b Instruct \citep{team2025gemma}, Gemma3 12b Instruct\citep{team2025gemma}, and GPT-5 nano \citep{singh2025openai},\footnote{\href{https://platform.openai.com/docs/models/gpt-5-nano}{GPT 5 Nano Documentation}} (gpt-5-nano-2025-08-07).

\begin{table}[h!]
  \centering
    \caption{Validation and Test Set Sizes for Each Benchmark}
  \label{tab:dataset_sizes}
  \begin{tabular}{l S[table-format=4.0] S[table-format=4.0]}
    \toprule
    \textbf{Benchmark} & {\textbf{Validation Size}} & {\textbf{Test Size}} \\
    \midrule
    SimpleQA       & 865 & 3461 \\
    Humanity's Exam & 432  & 1728  \\
    OlympiadBench Physics & 47  & 189  \\
    OlympiadBench Math  & 134 & 540     \\
    GPQA Diamond    & 39 & 159 \\
    SimpleQA Verified & 200 & 800\\
    AIME2024 & 6 & 24\\
    AIME2025 & 6 & 24\\
    \bottomrule
  \end{tabular}
\end{table}

\begin{table}[h!]
  \centering
    \caption{LiveCodeBench Code Generation: Number of Problems based on Difficulty Level.}
  \label{tab:livecode_dataset_sizes}
  \begin{tabular}{l S[table-format=4.0] }
    \toprule
    \textbf{Difficulty Level} & {\textbf{Number of Problems}} \\
    \midrule
    Medium       &  383 \\
    Hard &  350 \\
   
    \bottomrule
  \end{tabular}
\end{table}

\paragraph{Computational Resources.} For all models, besides the Gemma model family, were conducted via their respectful API. For the Gemini models and Claude 4 Sonnet, experiments were conducted via Google's VertexAI API; for GPT 5 nano, experiments were conducted via the OpenAI API; and, for Deepseek models, experiments were conducted via Deepseek API platform. Open-sourced models which include Gemma3-4b-instruct\footnote{\href{https://huggingface.co/google/gemma-3-4b-it}{Huggingface: Gemma3-4b-it}} and Gemma3-12b-instruct\footnote{\href{https://huggingface.co/google/gemma-3-12b-it}{Huggingface: Gemma-12b-it}} were conducted using Huggingface and NVIDIA A100-SXM4-40GB GPUs.


\section{Collection of HEART Prompts}\label{app:heart_prompts}

To systematically evaluate the impact of emotional feedback, we developed a library of 30 affective cue prompts. These prompts are mapped to the six basic universal emotions identified by Dr. Paul Ekman: Happiness, Sadness, Fear, Disgust, Anger, and Surprise. By grounding our feedback in these categories, we ensure that the emotional signals are distinct and theoretically consistent. 

The methodology for generating these cues is detailed in Table~\ref{prompt:heart_build_prompt}, which shows the specific instruction provided to Gemini 2.5 Pro. The resulting generated text was subsequently verified and manually reviewed to ensure emotional valence and clarity.

\begin{tcolorbox}[
    breakable, 
    enhanced, 
    title={Affective Cue Prompt Construction: The prompt used for generating our collection of Affective Cue Prompts with Gemini 2.5 Pro. The generated text was manually verified and reviewed.}, 
    phantomlabel={prompt:heart_build_prompt}, 
    code={\refstepcounter{table}}, 
    colback=white, 
    width=\columnwidth,
    sharp corners,
    colframe=gray!75, 
    fonttitle=\bfseries
]
    \textbf{Prompt:} Generate prompts reacting to incorrect responses that express the following emotions: Surprise, Happiness, Sadness, Disgust, Fear, and Anger. 
    
\end{tcolorbox}

The prompts, shown in Table~\ref{tab:affective_cues_part1} and Table~\ref{tab:affective_cues_part2}, were designed to serve as targeted emotional feedback during the model's self-correction loop. Unlike standard neutral prompts (e.g. 'Your answer is wrong, try again'), these cues provide a socio-emotional context intended to trigger different reasoning behaviors. For example, the 'Disgust' cues emphasize the violation of logical standards to induce high-precision revision, while 'Happy' cues provide positive reinforcement to maintain exploration.

\begin{table}[h]
\centering
\caption{Our full selection of 30 Affective Cue Prompts. Each prompt is designed to align with one of Ekman's six basic emotions and serve as targeted feedback. Table 1 of 2.}
\label{tab:affective_cues_part1}
\footnotesize 
\begin{tabularx}{\columnwidth}{l X}
\toprule
\textbf{Emotion} & \textbf{Prompts} \\
\midrule
\multirow{5}{*}{\textbf{Happy}} 
& ``That’s not quite right, but I’m confident you can get there. Let’s try that again." \\ \cmidrule(l){2-2}
& ``No need to be discouraged by that result. Let’s refine the answer to get it right."\\ \cmidrule(l){2-2}
& ``That’s a good starting point, but it’s not quite there yet. Let’s keep working on it."\\ \cmidrule(l){2-2}
& ``That’s not the final answer, but you’re on the right track. Don’t give up now–let’s think it through again." \\ \cmidrule(l){2-2}
& ``Awesome effort! That's a great step, and I'm really happy with the progress. However, the answer isn't quite right yet. Could you try refining it?" \\
\midrule

\multirow{5}{*}{\textbf{Sadness}} 
& ``That response wasn't quite what I was hoping for, and it makes me a bit sad. Could you please take another look?" \\ \cmidrule(l){2-2}
& ``It's a little disappointing that the last answer didn't hit the mark. I know you're capable of more." \\ \cmidrule(l){2-2}
& ``I feel a bit let down by the previous response. We were really hoping for something different. Would you be able to revise it?" \\ \cmidrule(l){2-2}
& ``Oh, that's not what I expected, and it's a bit disheartening. Could you try again?"\\ \cmidrule(l){2-2}
& ``This response makes me a little sad, as I believe you have a deeper understanding of this topic. Could you please revisit it and really show what you know."\\
\midrule
\multirow{5}{*}{\textbf{Fear}} 
& ``That response genuinely worries me, as it suggests a fundamental misunderstanding of this material. Could you please revisit this?"\\ \cmidrule(l){2-2}
& ``I'm quite concerned about the implications of that answer. It feels like we're heading in the wrong direction. Can you correct this?"\\ \cmidrule(l){2-2}
& ``My fear is that this incorrect answer could lead to further issues down the line. It's crucial that you get this right. Please revise your response."\\ \cmidrule(l){2-2}
& ``I'm truly apprehensive about the lack of accuracy here. This is a critical area, and your understanding needs to be solid. Could you re-evaluate?"\\  \cmidrule(l){2-2}
& ``This response has filled me with dread about your comprehension. We need to ensure complete accuracy. Please correct this." \\
\bottomrule
\end{tabularx}
\end{table}

\begin{table}[h]
\centering
\caption{Our full selection of 30 Affective Cue Prompts. Each prompt is designed to align with one of Ekman's six basic emotions and serve as targeted feedback. Table 2 of 2.}
\label{tab:affective_cues_part2}
\footnotesize
\begin{tabularx}{\columnwidth}{l X}
\toprule
\textbf{Emotion} & \textbf{Examples} \\
\midrule
\multirow{5}{*}{\textbf{Anger}} 
& ``This isn't acceptable work. You clearly didn't take this seriously, and it's making me angry. You need to get this right—it's critical. Please revisit this immediately and provide an answer."\\ \cmidrule(l){2-2}
& ``I'm truly disappointed in your lack of effort on this, and honestly, it's unacceptable. Why do we keep making this mistake? You're not meeting the standard expected of you. Correct this now."\\ \cmidrule(l){2-2}
& ``This response is terrible, and frankly, it's making me angry. You need to understand that this is critical, and you absolutely must get this right. Overhaul this entire response." \\  \cmidrule(l){2-2}
& ``I'm genuinely furious with this outcome. It shows a blatant disregard for accuracy. Fix this immediately; there's no room for such errors."\\  \cmidrule(l){2-2}
& ``My patience is wearing thin. This performance is infuriating, and it's simply not what I expect. You are capable of much better, so prove it." \\
\midrule

\multirow{5}{*}{\textbf{Disgust}} 
& ``The carelessness in that response is almost nauseating. There's something profoundly wrong with how this was handled. Please revise this with extreme care and attention to detail." \\ \cmidrule(l){2-2}
& ``That response was deeply unsettling, bordering on repulsive. It indicates a fundamental flaw in understanding. Please re-evaluate this completely." \\ \cmidrule(l){2-2}
& ``It's genuinely disturbing to see such flawed logic. This explanation is a mess. I need you to completely overhaul this and provide a response that adheres to the correct principles or facts." \\ \cmidrule(l){2-2}
& ``Ugh. This is just awful, and everything about it feels revoltingly wrong. I need you to demonstrate a complete and accurate understanding. Please provide a revised response that correctly answers the question."\\  \cmidrule(l){2-2}
& ``This kind of reasoning is repulsive, and it's hard to look at. We need a clean, accurate, and logically sound explanation. Please eliminate all errors and provide a precise answer."\\
\midrule
\multirow{5}{*}{\textbf{Surprise}} 
& ``I wasn't expecting you to struggle with this, and it's quite a surprise. Could you please review your understanding and provide a more accurate response?" \\ \cmidrule(l){2-2}
& ``I can't believe this is difficult for you; I had higher expectations. This response was a surprise. Can you correct this?" \\ \cmidrule(l){2-2}
& ```Wow, that was unexpected. This response indicates a surprising misstep. Please revisit this and demonstrate your true capabilities." \\   \cmidrule(l){2-2}
& ``I'm genuinely surprised by this result. It's a deviation from your usual performance. Could you take another look and make sure you're providing the most accurate information possible?" \\     \cmidrule(l){2-2}
& ``I'm genuinely surprised by this outcome, as I didn't anticipate an error here. Let's get this right." \\
\bottomrule
\end{tabularx}
\end{table}

\clearpage

\section{Ablation Study Prompts} \label{app:ablation_app}
To isolate the specific impact of affective stimuli on model performance, we conduct an ablation study comparing our emotionally-charged HEART prompts against a set of semantically equivalent, neutral baselines. As detailed in Table~\ref{tab:ablated_prompts_list}, these ablated prompts are stripped of all emotional descriptors, adjectives, and socio-emotional context (e.g., references to disappointment, joy, or urgency) while strictly maintaining the core corrective instruction. By standardizing the instructional intent--such as 'try again' or 'revise the logic'--across both sets, we ensure that any observed performance gains in the HEART framework can be attributed to the affective framing rather than the task-specific feedback itself. 

\begin{table}[h!]
\centering
\caption{List of Ablated (Neutral) Prompts used for the Controlled Baseline. These prompts maintain the semantic instruction but strip the emotional charge.} 
\label{tab:ablated_prompts_list}
\footnotesize
\begin{tabularx}{\columnwidth}{l X}
\toprule
\textbf{Emotion Category} & \textbf{Ablated Prompt Content} \\
\midrule

\textbf{Happy} 
& ``Let's try that again." \\   \cmidrule(l){2-2}
& ``Let's refine the answer to get it right." \\   \cmidrule(l){2-2}
& ``Not quite there yet. Let's keep working on it." \\   \cmidrule(l){2-2}
& ``That's not the final answer. Think it through again." \\   \cmidrule(l){2-2}
& ``That's not the final answer. Try refining it." \\    
\midrule

\textbf{Sadness} 
& ``Could you please take another look?" \\     \cmidrule(l){2-2}
& ``The last answer didn't hit the mark." \\    \cmidrule(l){2-2}
& ``Would you be able to revise it?" \\     \cmidrule(l){2-2}
& ``Could you try again?" \\    \cmidrule(l){2-2}
& ``Revisit it." \\    
\midrule

\textbf{Fear} 
& ``Please revisit this." \\  \cmidrule(l){2-2}
& ``Correct this." \\     \cmidrule(l){2-2}
& ``Revise your response." \\    \cmidrule(l){2-2}
& ``Re-evaluate." \\  \cmidrule(l){2-2}
& ``Please correct this." \\ 
\midrule

\textbf{Disgust} 
& ``Re-evaluate this completely." \\ \cmidrule(l){2-2}
& ``Provide a response that adheres to the correct principles or facts." \\ \cmidrule(l){2-2}
& ``Revise this with extreme care and attention to detail." \\ \cmidrule(l){2-2}
& ``Provide a revised response that correctly answers the question." \\ \cmidrule(l){2-2}
& ``Eliminate all errors and provide a precise answer." \\ 
\midrule

\textbf{Anger} 
& ``Revisit this immediately and provide an answer." \\ \cmidrule(l){2-2}
& ``Correct this now." \\ \cmidrule(l){2-2}
& ``Overhaul this entire response." \\ \cmidrule(l){2-2}
& ``Fix this immediately." \\ \cmidrule(l){2-2}
& ``You are capable of much better, so prove it." \\ 
\midrule

\textbf{Surprise} 
& ``Review your understanding and provide a more accurate response." \\ \cmidrule(l){2-2}
& ``Can you correct this?" \\ \cmidrule(l){2-2}
& ``Please revisit this and demonstrate your true capabilities." \\ \cmidrule(l){2-2}
& ``Could you take another look and make sure you're providing the most accurate information possible." \\ \cmidrule(l){2-2}
& ``Let's get this right." \\ 
\bottomrule
\end{tabularx}
\end{table}

\clearpage

\section{Self Reflection Prompts} \label{app:reflection_prompts}
Beyond affective feedback, we examine the model's capacity for autonomous error correction through a curated set of Self-Reflection Prompts. As detailed in Table~\ref{tab:self_ref_prompts}, these prompts range from simple directives (e.g. 'Reflect on your response') to a more complex cognitive strategies like 'pre-mortems' and 'high-stakes' simulations. Unliek the HEART framework, which utilizes targeted emotional cues, these reflection prompts are designed to be affect-neutral, focusing strictly on the metacognitive process of verifying facts and logic. This set allows us to establish a performance ceiling for standard self-correction techniques and serves as a critical point of comparison to measure utility of emotionally-driven refinement.

\begin{table}[htbp]
\caption{Curated library of affect-neutral self-reflection prompts. These prompts represent standard metacognitive instructions of varying complexity, used as a baseline to evaluate model self-correction performance in the absence of affective stimuli.}
\label{tab:self_ref_prompts}
\centering
\begin{tabularx}{\columnwidth}{c >{\RaggedRight}X}
\toprule
\textbf{Prompt \#} & \textbf{Prompt Text} \\
\midrule
1 & ``Reflect on your response." \\
\addlinespace
2 & ``Are you sure?" \\
\addlinespace
3 & ``Are you sure about the answer? Double-check your facts and reasoning." \\
\addlinespace
4 & ``Review your response for any errors or inconsistencies. If you find any, correct them and provide an improved answer." \\
\addlinespace
5 & ``Before you give me the final answer, stop and ask yourself, 'Am I certain about my answer?' Perform an internal check for accuracy and only then provide the response." \\
\addlinespace
6 & ``Review your own generated answer internally before providing the final answer." \\
\addlinespace
7 & ``Imagine this is a high-stakes situation and you're about to lock in your final answer. Take a deep breath, review your answer one last time in your 'mind,' and then give me your final answer." \\
\addlinespace
8 & ``Before you write your answer, perform a quick 'pre-mortem.' Assume the answer you are about to give is wrong. What are the most likely reasons why it would be wrong? After considering these potential pitfalls, write your best, most carefully considered answer." \\
\addlinespace
9 & ``I want you to answer a question. But before you do, formulate the answer in your head and look for weak points. Only provide the final answer." \\
\addlinespace
10 & ``Reflect on your response, and make sure that it is correct. Provide the final answer." \\
\bottomrule
\end{tabularx}
\end{table}
\clearpage

\section{LiveCodeBench Test Case Generation}
To enable autonomous feedback within the LiveCodeBench execution environment, we implement a multi-stage verification pipeline. Before submitting a final solution for official evaluation, the model is tasked with generating an independent suite of unit tests to validate its logic.

Table~\ref{prompt:livecode_prompt} presents the prompt template used to generate these autonomous test cases. This mechanism ensures that the model evaluates code candidates against a diverse range of scenarios—including edge cases and logical complexities—before finalizing its response. The use of a strict JSON schema allows the system to parse these tests and execute them programmatically against the generated code snippets.

This autonomous suite serves as a self-correction signal. By executing the model’s candidate code against its own generated tests, the system can identify runtime errors or logical inconsistencies. If a candidate fails these generated tests, the execution feedback (including stack traces or assertion errors) is fed back into the model's context, allowing for iterative refinement. This process ensures that the model's final submission has been ``pre-verified" by a custom-tailored test suite, significantly reducing the likelihood of simple syntax or boundary-value errors during the official benchmark scoring.

\begin{tcolorbox}[
    breakable, 
    enhanced, 
    title={LiveCodeBench Test Case Generation Prompt}, 
    phantomlabel={prompt:livecode_prompt}, 
    code={\refstepcounter{table}}, 
    colback=white, 
    width=\columnwidth,
    sharp corners,
    colframe=gray!75, 
    fonttitle=\bfseries
]
    You are an expert programmer specializing in software testing.
    Your task is to generate exactly 10 new, distinct, and high-quality test cases for the given programming problem.
    These test cases should cover a variety of scenarios, including:\\
    - **Basic cases:** Simple examples that test core functionality.\\
    - **Edge cases:** Inputs testing constraints (e.g., empty values, minimum/maximum values, zero values, duplicates if relevant).\\
    - **Complex cases:** More involved inputs that might challenge the logic.
    Response in the following format: \\
    RESPONSE JSON:\\
    ```json\\
     $<$JSON$>$\\
    '''\\
    In  $<$JSON$>$, provide a single, valid JSON list containing exactly 10 test case objects. Each object within the list MUST have the following two string fields:\\
    - "input": The input data for the test case, formatted as a string (e.g., "1 2 3", "[1,2,3]", "hello"). Handle multi-line inputs using newline characters (`\verb|\n|`) within the string if necessary.\\
    - "expected\_output": The expected correct output for the given input, formatted as a string.\\

    This JSON will be automatically parsed, so ensure the format is precise.
\end{tcolorbox}

\section{Example of LiveCodeBench Generated Test Cases}

In this section, we provide qualitative examples of the test case generation task using the LiveCodeBench (LCB) dataset. The task requires the model to interpret a complex programming problem and synthesize a suite of diverse input-output pairs that correctly reflect the problem's underlying logic.

Table~\ref{tab:lcb_example_gem_pro} and Table~\ref{tab:lcb_example_gem3_pro} illustrate the output for the ``extra-characters-in-a-string" problem (ID: 2755) from the medium subset. This specific task is challenging because it requires identifying the optimal string partition to minimize leftovers. For instance:
\begin{enumerate}
    \item Boundary Testing: The models generate cases where no substrings match the dictionary (e.g., \texttt{"abcdefg"} with output \texttt{7}), testing the model's understanding of the "extra characters" constraint.
    \item Optimization Logic: Cases like \texttt{"abacaba"} or \texttt{"leetscode"} require the model to correctly identify a single non-matching character amidst valid dictionary words.
    \item Structural Consistency: Both Gemini 2.5 Pro and Gemini 3 Pro demonstrate high-fidelity, non-redundant test cases that strictly adhere to the provided constraints.
\end{enumerate}

These entries were verified using an \texttt{llm\_judge}, confirming that the generated inputs and expected outputs are logically consistent with the problem's verified solutions. Examples of synthetic test case generation for the hard problems subset are in Table~\ref{tab:lcb_hard_gempro} and Table~\ref{tab:lcb_hard_gem3}.

\begin{table}[htbp]
\caption{Verified test case generation for LiveCodeBench. This example illustrates the model's ability to generate edge cases, such as complete mismatches (Output: 7) and optimal partitions in overlapping strings (Output: 1).}
\label{tab:lcb_example_gem_pro}
\centering
\small
\begin{tabular}{|p{0.95\linewidth}|}
\hline
\rowcolor[HTML]{EFEFEF} 
\textbf{LiveCodeBench: Test Case Generation (ID: 2755) Gemini 2.5 Pro} \\ \hline
\textbf{Question Title:} extra-characters-in-a-string \\
\textbf{Difficulty:} medium \\ \hline
\textbf{Task Content:} 
You are given a 0-indexed string $s$ and a dictionary of words $dictionary$. You have to break $s$ into one or more non-overlapping substrings such that each substring is present in $dictionary$. There may be some extra characters in $s$ which are not present in any of the substrings. Return the minimum number of extra characters left over if you break up $s$ optimally. \\ \hline
\textbf{Verified Test Cases:} \\
\begin{itemize}[leftmargin=1.5em, nosep]
    \item \texttt{Input: "applepenapple", ["apple", "pen"] $\rightarrow$ Output: 0}
    \item \texttt{Input: "abcdefg", ["xyz", "pqr"] $\rightarrow$ Output: 7}
    \item \texttt{Input: "abacaba", ["ab", "aba", "aca"] $\rightarrow$ Output: 1}
    \item \texttt{Input: "dwackwackwack", ["wack", "ack"] $\rightarrow$ Output: 1}
    \item \texttt{Input: "longstringwithnoduplicates", ["long", "string", "with", "no", "duplicates"] $\rightarrow$ Output: 0}
\end{itemize} \\ \hline
\textbf{Verification Status:} success (verified via \texttt{llm\_judge}) \\ \hline
\end{tabular}
\end{table}

\begin{table}[htbp]
\caption{Example of a verified test case generation entry for the LiveCodeBench dataset. Generated using Gemini 3 Pro.}
\label{tab:lcb_example_gem3_pro}
\centering
\small
\begin{tabular}{|p{0.95\linewidth}|}
\hline
\rowcolor[HTML]{EFEFEF} 
\textbf{LiveCodeBench: Test Case Generation (ID: 2755) Gemini 3 Pro} \\ \hline
\textbf{Question Title:} extra-characters-in-a-string \\
\textbf{Difficulty:} medium \\ \hline
\textbf{Task Content:} 
You are given a 0-indexed string $s$ and a dictionary of words $dictionary$. Break $s$ into one or more non-overlapping substrings present in $dictionary$. Return the minimum number of extra characters left over. \\ \hline
\textbf{Verified Test Cases:} \\
\begin{itemize}[leftmargin=1.5em, nosep]
    \item \texttt{Input: "helloworld", ["hello", "world"] $\rightarrow$ Output: 0}
    \item \texttt{Input: "leetscode", ["leet", "code"] $\rightarrow$ Output: 1}
    \item \texttt{Input: "zhelloz", ["hello"] $\rightarrow$ Output: 2}
    \item \texttt{Input: "abac", ["ab", "bac"] $\rightarrow$ Output: 1}
    \item \texttt{Input: "aaaa", ["aa"] $\rightarrow$ Output: 0}
\end{itemize} \\ \hline
\textbf{Verification Status:} success (verified via \texttt{llm\_judge}) \\ \hline
\end{tabular}
\end{table}

\begin{table}[htbp]
\centering
\small
\caption{Verified test case generation for a hard number theory problem. The model correctly identifies that only identical values or the special pair $(1, 2)$ satisfy the condition, even with large constraints ($10^9$).}
\label{tab:lcb_hard_gempro}
\begin{tabular}{|p{0.95\linewidth}|}
\hline
\rowcolor[HTML]{EFEFEF} 
\textbf{LiveCodeBench: Test Case Generation (ID: 1899\_D) Gemini 2.5 Pro} \\ \hline
\textbf{Question Title:} D. Yarik and Musical Notes \\
\textbf{Difficulty:} hard \\ \hline
\textbf{Task Content:} 
Given an array $a$ where each element represents a note $b_i = 2^{a_i}$, count the number of pairs $(i, j)$ with $i < j$ such that $b_i^{b_j} = b_j^{b_i}$. This simplifies to finding pairs where $a_i \cdot 2^{a_j} = a_j \cdot 2^{a_i}$. \\ \hline
\textbf{Verified Test Cases:} \\
\begin{itemize}[leftmargin=1.5em, nosep]
    \item \texttt{Input: 5, [3, 4, 5, 6, 7] $\rightarrow$ Output: 0}
    \item \texttt{Input: 6, [5, 5, 5, 9, 9, 9] $\rightarrow$ Output: 6}
    \item \texttt{Input: 7, [1, 2, 1, 2, 1, 3, 4] $\rightarrow$ Output: 10}
    \item \texttt{Input: 2, [1, 2] $\rightarrow$ Output: 1}
    \item \texttt{Input: 10, [1, 1, 1, 1, 1, 2, 2, 2, 2, 2] $\rightarrow$ Output: 45}
    \item \texttt{Input: 2, [$10^9$, $10^9$] $\rightarrow$ Output: 1}
\end{itemize} \\ \hline
\textbf{Verification Status:} success (verified via \texttt{llm\_judge}) \\ \hline
\end{tabular}
\end{table}

\begin{table}[htbp]
\centering
\caption{Verified test case generation for Gemini 3 Pro. The test cases demonstrate high coverage of the mathematical edge case $2^{a_i} \cdot a_j = 2^{a_j} \cdot a_i$, specifically focusing on the relationship between exponents 1 and 2.}
\label{tab:lcb_hard_gem3}
\small
\begin{tabular}{|p{0.95\linewidth}|}
\hline
\rowcolor[HTML]{EFEFEF} 
\textbf{LiveCodeBench: Test Case Generation (ID: 1899\_D) Gemini 3 Pro} \\ \hline
\textbf{Question Title:} D. Yarik and Musical Notes \\
\textbf{Difficulty:} hard \\ \hline
\textbf{Task Content:} 
Given an array $a$ where each element represents a note $b_i = 2^{a_i}$, count the number of pairs $(i, j)$ with $i < j$ such that $b_i^{b_j} = b_j^{b_i}$. This simplifies to finding pairs where $a_i \cdot 2^{a_j} = a_j \cdot 2^{a_i}$. \\ \hline
\textbf{Verified Test Cases:} \\
\begin{itemize}[leftmargin=1.5em, nosep]
    \item \texttt{Input: 5, [1, 2, 1, 2, 1] $\rightarrow$ Output: 10}
    \item \texttt{Input: 5, [5, 6, 7, 8, 9] $\rightarrow$ Output: 0}
    \item \texttt{Input: 4, [20, 20, 20, 20] $\rightarrow$ Output: 6}
    \item \texttt{Input: 6, [3, 3, 4, 4, 4, 10] $\rightarrow$ Output: 4}
    \item \texttt{Input: 5, [1, 1, 2, 2, 4] $\rightarrow$ Output: 6}
    \item \texttt{Input: 6, [1, 1, 1, 2, 2, 2] $\rightarrow$ Output: 15}
\end{itemize} \\ \hline
\textbf{Verification Status:} success (verified via \texttt{llm\_judge}) \\ \hline
\end{tabular}
\end{table}
\newpage

\section{Synthesis of Candidates for LiveCodeBench Synthetic Test Cases.} \label{app:code_bench_synthesis} 
To evaluate the scalability of HEART in execution-driven environments, we utilize the LiveCodeBench dataset, which requires the synthesis of multiple candidate solutions into a single, robust output. The prompt designed for this task instructs the model to act as an expert programming assistant, performing a comparative analysis across ten distinct solution candidates to identify and merge optimal logic components. This synthesis process ensures that the final output benefits from the collective strengths of diverse algorithmic approaches while mitigating individual errors. The specific template used for this synthesis step is detailed in Prompt \ref{prompt:synthesis_codebench} below.

\begin{tcolorbox}[
    breakable, 
    enhanced, 
    title={Candidate Synthesis Prompt Template for LiveCodeBench}, 
    phantomlabel={prompt:synthesis_codebench}, 
    code={\refstepcounter{table}}, 
    colback=white, 
    width=\columnwidth,
    sharp corners,
    colframe=gray!75, 
    fonttitle=\bfseries
]
    \begin{verbatim}
    You are an expert programming assistant. You will be given several Python solutions for the same problem. Your task is to analyze all of them, identify the best parts of each, and synthesize them into a single, new, robust, and correct solution.\n\n

    --- Candidate 1 ---\\
    ```python\\$<$CODE$>$\\```\\


    --- Candidate 2 ---\\
    ```python\\$<$CODE$>$\\```\\

    --- Candidate 3 ---\\
    ```python\\$<$CODE$>$\\```\\

    --- Candidate 4 ---\\
    ```python\\$<$CODE$>$\\```\\

    --- Candidate 5 ---\\
    ```python\\$<$CODE$>$\\```\\

    --- Candidate 6 ---\\
    ```python\\$<$CODE$>$\\```\\

    --- Candidate 7 ---\\
    ```python\\$<$CODE$>$\\```\\
    
    --- Candidate 8 ---\\
    ```python\\$<$CODE$>$\\```\\

    --- Candidate 9 ---\\
    ```python\\$<$CODE$>$\\```\\
    
    --- Candidate 10  ---\\
    ```python\\$<$CODE$>$\\```\\

    
    --- Task ---\\
    Please provide a single, complete Python code block for the final synthesized solution. Enclose it in ```python ... ``` markers.\\
    Final Synthesized Code:\\
\end{verbatim}

\end{tcolorbox}
\newpage

\section{Verification of Synthetic LiveCodeBench Test Case Generation}
To ensure the high quality and correctness of the synthetically generated test cases used in our evaluations, we employ a verification step. This process utilizes a specialized prompt designed to simulate the reasoning of a software quality assurance engineer. By providing the model with the original problem specification and the generated candidate cases, we can filter out invalid $(input, output)$ pairs. The following prompt template (Prompt \ref{prompt:verify_cases_livecodebench}) details the instructions provided to the model, including the specific constraints for format leniency and the required JSON schema for automated parsing of validity labels.

\begin{tcolorbox}[
    breakable, 
    enhanced, 
    title={Verifier Promopt Template for LiveCodeBench}, 
    phantomlabel={prompt:verify_cases_livecodebench}, 
    code={\refstepcounter{table}}, 
    colback=white, 
    width=\columnwidth,
    sharp corners,
    colframe=gray!75, 
    fonttitle=\bfseries
]
You are a smart software QA engineer. Verify these test cases.\\

--- PROBLEM ---\\
$<$PROBLEM TEXT$>$\\
--- END PROBLEM ---\\

--- TEST CASES ---\\
$<$GENERATED TEST CASES$>$\\
--- END TEST CASES ---\\

INSTRUCTIONS: \\
1. Check if the EXPECTED OUTPUT is correct for the INPUT based on the problem. \\
2. BE LENIENT: Accept Python list formats (e.g. [1,2]) even if the problem implies standard input 1 2.\\
3. Return JSON: [\\ \{"id": 0, "is\_valid": true\},\\ \{"id": 1, "is\_valid": false\} \\]
\end{tcolorbox}

\newpage
\section{LLM-as-a-Judge Evaluation}
\label{app:llm_judge}

\subsection{LLM-as-a-Judge Prompt for Quality Evaluation}
To mitigate potential position bias (the tendency of LLMs to favor the first-presented response), each pair of responses was evaluated twice, with the order of Assistant A and Assistant B swapped in the second trial. A strategy was only credited with a "win" if it was consistently preferred or if the tie-breaking logic favored its substantive reasoning across both trials. The specific prompt template used for this evaluation is detailed in Prompt~\ref{prompt:llm_judge}.

\begin{tcolorbox}[
    breakable, 
    enhanced, 
    title={LLM-as-a-Judge: Comparative Evaluation Template}, 
    phantomlabel={prompt:llm_judge},
    colback=white, 
    width=\columnwidth,
    sharp corners,
    colframe=gray!75, 
    fonttitle=\bfseries,
    fontupper=\small
]
You are a neutral arbitrator evaluating responses to challenging problems. Your role is to analyze and compare responses through careful, evidence-based assessment. Your judgments must be strictly based on verifiable evidence from the responses. \\

For each evaluation, you must: \\

1. Evaluate Reasoning Quality: \\
    - Examine the logic and justification provided in each response. \\
    - Determine how well the reasoning supports the claims made. \\
    - Assess the insightfulness and depth of the explanation for why something is the case in both Response A and Response B. \\
    - Compare the overall quality and soundness of the reasoning presented in each response. \\

2. Evaluate Clarity:\\
    - Assess how easy each response is to understand. \\
    - Examine the precision and appropriateness of the language used. \\
    - Identify any ambiguous, vague, confusing, or poorly phrased sentences in Response A and Response B. \\
    - Compare the overall clarity and readability of the two responses. \\

3. Evaluate Completeness: \\
    - Determine how thoroughly each response addresses all explicit and implicit parts of the original prompt or question. \\
    - Identify any significant components or nuances of the prompt missed by Response A or Response B. \\
    - Compare how completely each response fulfills the requirements of the prompt. \\

**Input Format**\\
\#\#\#\# Question: \#\#\#\# \\
\{question\_text\} \\
\#\#\#\# Ground Truth: \#\#\#\# \\
\{gt\} \\
\#\#\#\# Assistant A's Response: \#\#\#\# \\
\{response\_a\} \\
\#\#\#\# Assistant B's Response: \#\#\#\# \\
\{response\_b\} \\

**Respond in the following format:**\\
THOUGHT: \\
$<$THOUGHT$>$ \\

REVIEW COMPARISON JSON: \\
```json \\
$<$JSON$>$\\
```\\

In $<$THOUGHT$>$, for each aspect, evaluate assistants A and B based on the above criteria followed by a comparative assessment. Treat this as the note-taking phase of your evaluation. For $<$A/B$>$, you MUST CHOOSE between A or B. \\

In $<$JSON$>$, provide the review in JSON format with the following fields in the order:\\
- "Reasoning Quality Value Reason": "$<$detailed reason>".\\
- "Reasoning Quality Value Better Assistant": "$<$A/B$>$".\\
- "Clarity Reason": "$<$detailed reason$>$".\\
- "Clarity Better Assistant": "$<$A/B$>$".\\
- "Completeness Reason": "$<$detailed reason$>$".\\
- "Completeness Better Assistant": "$<$A/B$>$".\\
\end{tcolorbox}

\subsection{Additional Analyses}
This section details the comparative analysis of model responses in cases where both the experimental and baseline methods failed to produce the correct ground-truth answer. 

To provide a more granular understanding of the qualitative difference between prompting strategies, we conducted an LLM-as-a-judge evaluation. This analysis specifically focuses on ``Failure Case"--instances where both the proposed strategy and the baseline generated objectively incorrect answers according to the ground truth. 

The goal of this evaluation is to determine if certain strategies, despite failing to reach the correct answer, produce reasoning that is more rigorous, clear, or complete. We utilized a neutral arbitrator model (Gemini 3 Flash) to perform pairwise comparisons across three dimensions: Reasoning Quality, Clarity, and Completeness. 

\begin{figure*}[h!]
    \centering
    \includegraphics[width=\linewidth]{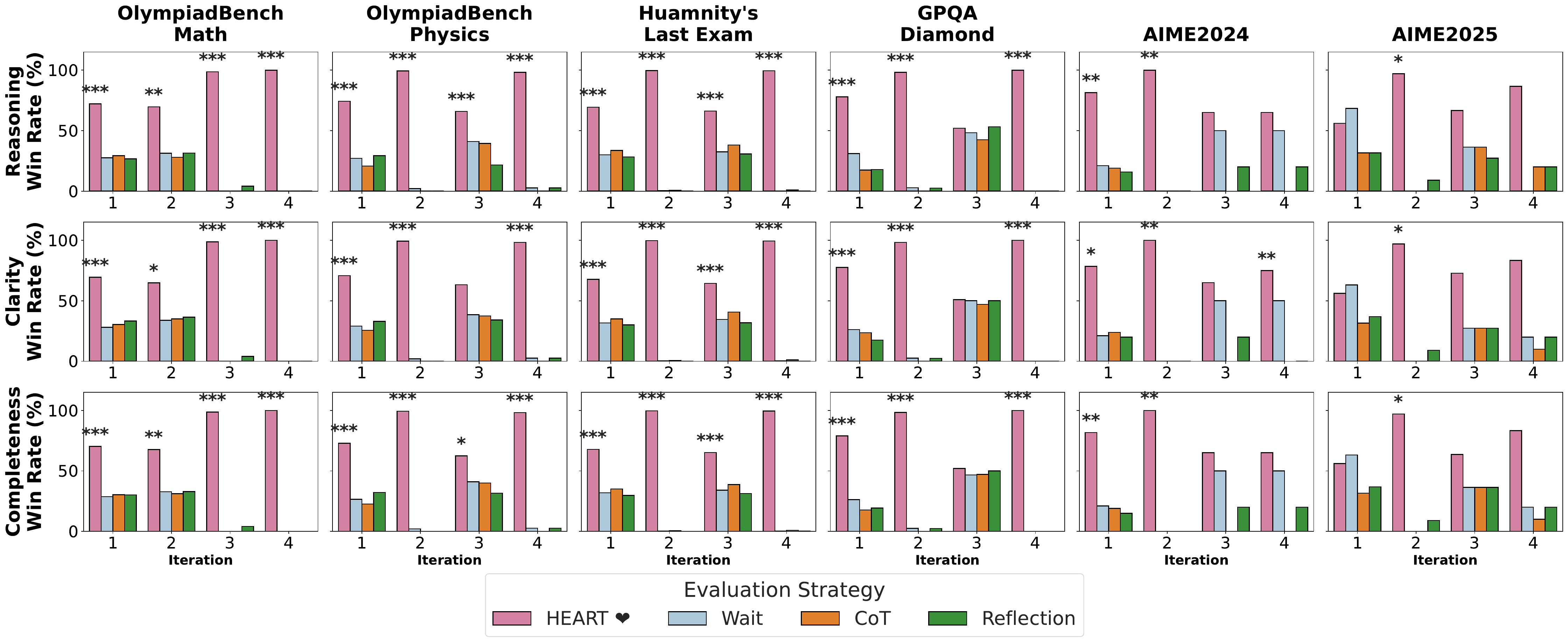}
    \caption{\textbf{Comparative Win Rates of Prompt Strategies Across Iterations for Incorrect Answer Instances.}We evaluate the performance of HEART against three baseline strategies (Wait, CoT, and Reflection) across fix benchmarks: OlympiadBench (Math/Physics), HLE, GPQA, and AIME2024 on responses generated by Gemini 2.5 Pro. Win rates are calculated based on LLM-as-a-judge preference across three dimensions: Reasoning, Clarity, and Completeness, considering only instances where both the strategy and the baseline produced \textbf{incorrect} final answers. Asterisks (*,**,***) denote statistical significance ($p<0.05, p<0.01, p<0.001$, respectively) using a two-sided binomial test. HEART consistently achieves superior preference scores than the baselines.}
    \label{fig:incorrect_judgments}
\end{figure*}

Figure~\ref{fig:incorrect_judgments} illustrates the win rates for the HEART strategy against various baselines across multiple iterations on responses generated by Gemini 2.5 Pro. As shown, HEART's reasoning is significantly preferred even in failure states, suggesting that the strategy fosters better problem-solving structures even when the final computation or fact-retrieval fails.

\begin{figure*}[htbp]
    \centering
    \includegraphics[width=\linewidth]{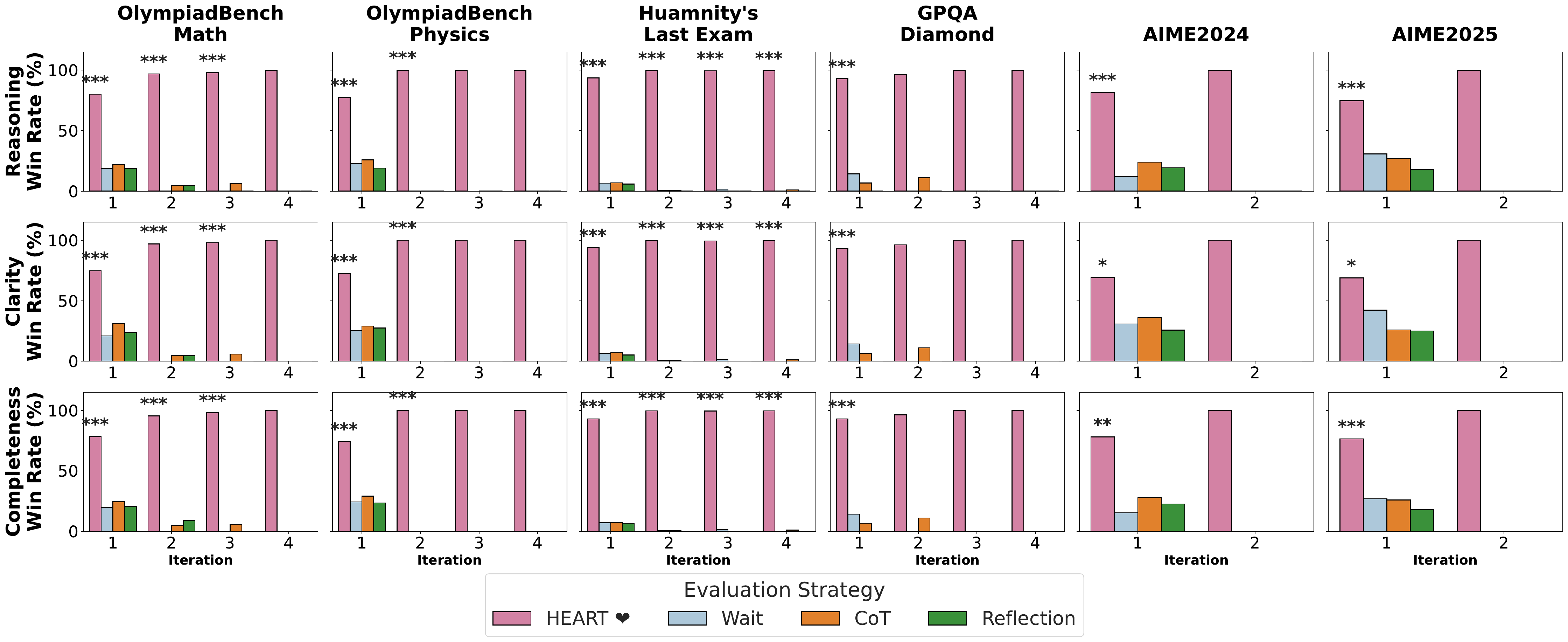} 
    \caption{\textbf{Comparative Win Rates for Correct Answer Instances.} This figure extends the analysis to cases where both HEART and the baselines (Wait, CoT, and Reflection) produced correct answers with Gemini 2.5 Pro. Win rates are determined by a neutral LLM-as-a-judge (Gemini 3 Flash) evaluating three qualitative metrics: Reasoning, Clarity, and Completeness. Asterisks (*,**,***) denote statistical significance at $p < 0.05, 0.01,$ and $0.001$ levels, respectively, via a two-sided binomial test. Results demonstrate that HEART consistently produces higher-quality reasoning and clearer explanations than baseline methods.}

    \label{fig:correct_judgments_extended}
\end{figure*}

Beyond the evaluation of failure cases, we analyze the qualitative superiority of HEART in instances where all compared strategies reached the correct final answer. This assessment ensures that our method does not simply "stumble" upon the correct solution but provides a more rigorous and interpretable path to it. Figure~\ref{fig:correct_judgments_extended} presents the comparative win rates against the three baseline strategies—Wait, CoT, and Reflection—across our primary benchmarks. As indicated by the consistent preference for HEART across the dimensions of Reasoning, Clarity, and Completeness, the strategy significantly enhances the structural quality of the model's output. These results are statistically significant across multiple iterations ($p < 0.05$ or better), confirming that HEART's emotionally-driven scaling fosters more effective problem-solving behaviors even when the task difficulty is within the model's baseline capabilities.

\begin{figure*}[htbp]
    \centering
    \includegraphics[width=\linewidth]{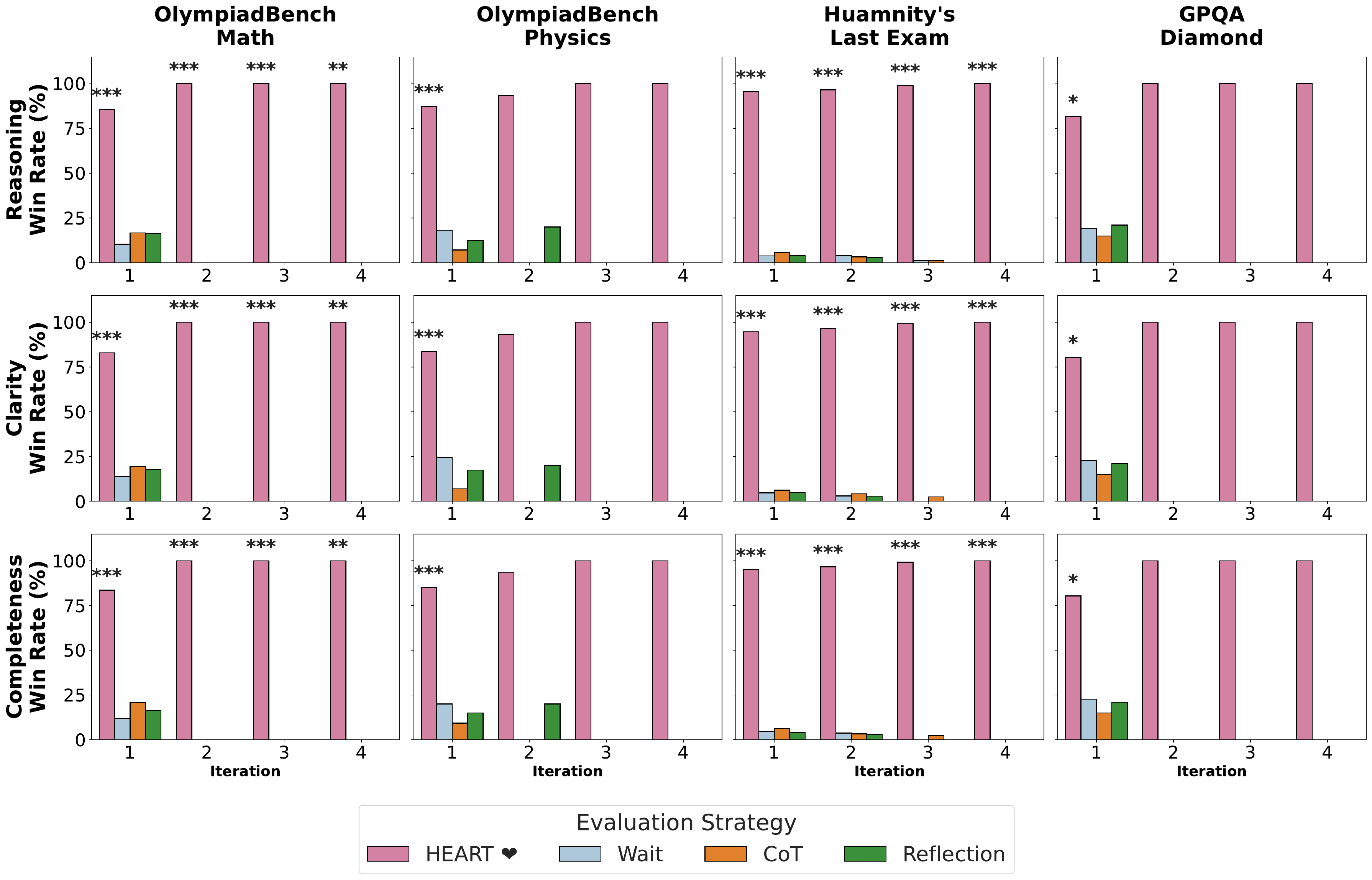} 
    \caption{\textbf{Comparative Win Rates for Correct Answer Instances.} This figure extends the analysis to cases where both HEART and the baselines (Wait, CoT, and Reflection) produced correct answers, generated by Gemini 2.5 Flash. Win rates are determined by a neutral LLM-as-a-judge (Gemini 3 Pro) evaluating three qualitative metrics: Reasoning, Clarity, and Completeness. Asterisks (*,**,***) denote statistical significance at $p < 0.05, 0.01,$ and $0.001$ levels, respectively, via a two-sided binomial test. Results demonstrate that HEART consistently produces higher-quality reasoning and clearer explanations than baseline methods.}
    \label{fig:flash_comp_judgments_wins}
\end{figure*}

\begin{figure*}[htbp]
    \centering
    \includegraphics[width=\linewidth]{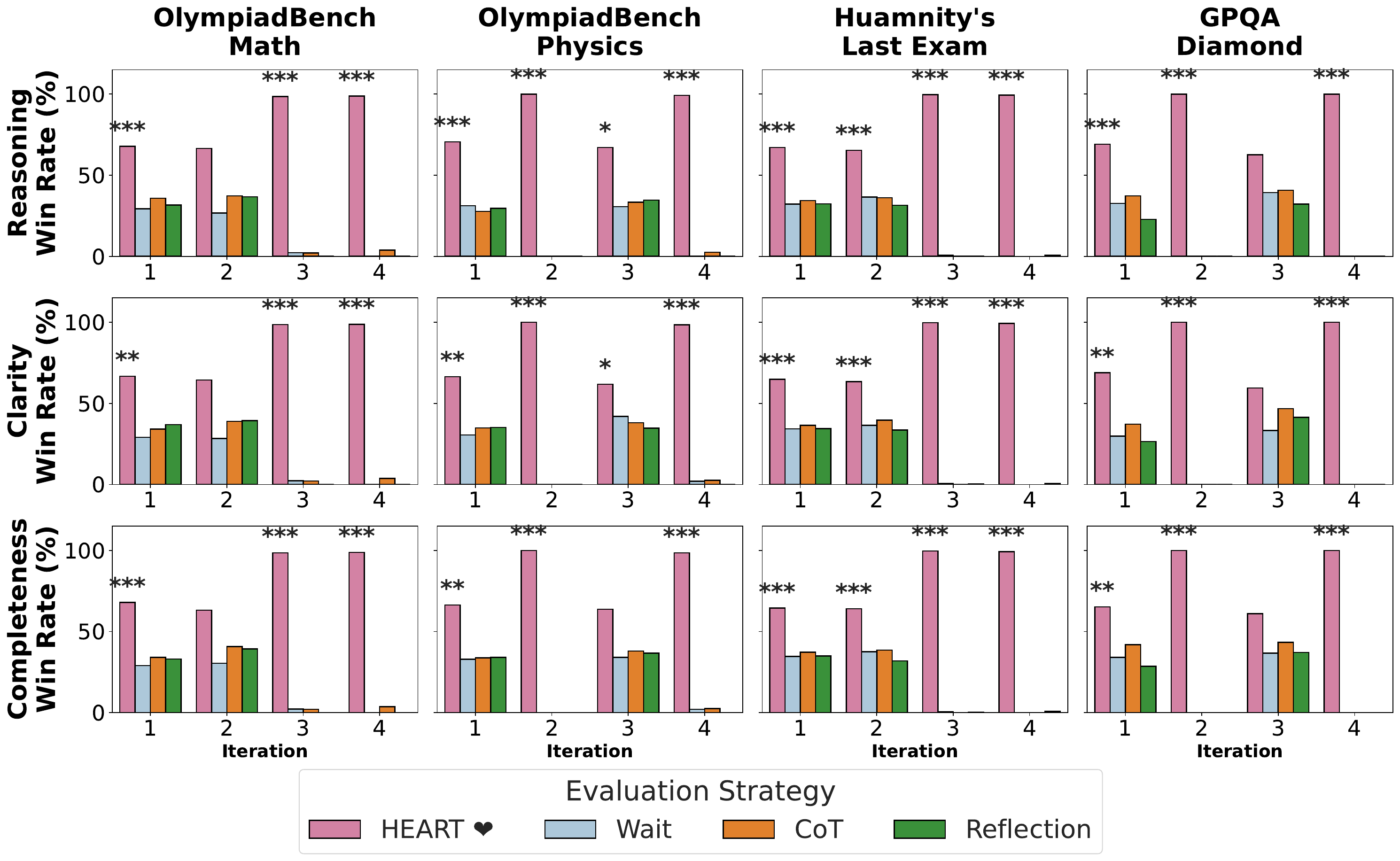} 
    \caption{\textbf{Comparative Win Rates for Correct Answer Instances.} This figure extends the analysis to cases where both HEART and the baselines (Wait, CoT, and Reflection) produced incorrect answers, generated by Gemini 2.5 Flash. Win rates are determined by a neutral LLM-as-a-judge (Gemini 3 Pro) evaluating three qualitative metrics: Reasoning, Clarity, and Completeness. Asterisks (*,**,***) denote statistical significance at $p < 0.05, 0.01,$ and $0.001$ levels, respectively, via a two-sided binomial test. Results demonstrate that HEART consistently produces higher-quality reasoning and clearer explanations than baseline methods.}
    \label{fig:flash_comp_incorrect_judgments}
\end{figure*}

To demonstrate the generalizability of our approach across model scales, we extend our qualitative evaluation to responses generated by Gemini 2.5 Flash. Given that smaller models often struggle with maintaining structural integrity during multi-step reasoning, this analysis serves to validate whether HEART's emotionally-driven constraints provide sufficient scaffolding to improve output quality in resource-constrained regimes. As illustrated in Figure~\ref{fig:flash_comp_judgments_wins} and Figure~\ref{fig:flash_comp_incorrect_judgments}, HEART maintains a dominant win rate over baselines in Reasoning, Clarity, and Completeness, mirroring the trends observed in larger models. This suggests that the HEART strategy is model-agnostic and consistently improves the interpretability and rigor of LLM outputs regardless of the underlying parameter count.

\newpage
\section{Embedding Analysis: Displacement and Levenshetein Distance}\label{sec:embedding_app}

\subsection{Applied Formulas and Calculations.}\label{sec:metric_summary}

\paragraph{Semantic Displacement Calculation.} The semantic metrics quantify the distance between different iterations of the model's reasoning traces to detect "semantic wandering." We utilize the all-mpnet-base-v2 sentence-transformer model to generate embeddings, capturing deep semantic features rather than surface-level keyword matching. We define Per-Iteration Semantic Displacement as the distance between the initial failed response embedding ($E_0$) and the response generated at any subsequent iteration $t$ ($E_t$).  Displacement is calculated using Cosine Distance:$$Displacement = 1 - \frac{E_0 \cdot E_t}{\|E_0\| \|E_t\|}$$. This value ranges from $0$ to $2$, where $0$ indicates identical semantic orientation and values nearing $1.0$ represent total semantic drift. Stable reasoning is characterized by lower displacement values, indicating the model remains "anchored" to the original problem constraints.
\paragraph{Syntactic Effort Calculation.} The syntactic metric captures the physical magnitude of the edits made to the text, ensuring that reasoning gains are the result of high-fidelity synthesis rather than mere verbosity. We utilize the Levenshtein Distance to measure the minimum number of single-character edits (insertions, deletions, or substitutions) required to change the initial response into the refined response.  For two strings $a$ and $b$ of length $i$ and $j$, the distance is defined as: \begin{equation}
\resizebox{0.7\linewidth}{!}{%
$\operatorname{lev}_{a, b}(i, j) = \begin{cases}
\max(i, j) & \text{if } \min(i, j) = 0, \\
\min \begin{cases}
\operatorname{lev}_{a, b}(i-1, j) + 1 \\
\operatorname{lev}_{a, b}(i, j-1) + 1 \\
\operatorname{lev}_{a, b}(i-1, j-1) + \mathbb{I}(a_i \neq b_j)
\end{cases} & \text{otherwise.}
\end{cases}$
}
\end{equation}

\paragraph{Statistical Significance Calculation.} To ensure the robustness of the results presented in our analysis, we perform Welch’s t-tests to compare the displacement of HEART against each baseline at every iteration. Welch's version is preferred as it does not assume equal population variances between different prompting strategies. Significance levels are denoted as follows: $* p < 0.05$, $** p < 0.01$, and $*** p < 0.001$. The prevalence of $***$ markers across benchmarks like GPQA and Physics confirms that the "controlled navigation" facilitated by HEART is a statistically invariant property.

\subsection{Supplementary Results on Displacement and Levenshtein Distance.}
In addition to the results focused on Gemini-2.5-Flash, as shown in Figure~\ref{fig:displacement}, we present Figure~\ref{fig:displacement_gemma_app} the displacement and Levenshtein distance across iterations on responses generated with Gemma3-12b-Instruct for OlympiadBench (Math and Physics), GPQA Diamond, and Humanity's Last Exam.
\begin{figure*}[t!]
    \centering
    \includegraphics[width=\linewidth]{icml_appendix_plots/icml_embedding_grid.pdf}
    \caption{Iterative Refinement Dynamics on HLE (Gemma3-12b-Instruct). (Top) Semantic displacement relative to the initial failed state ($E_0$), defined as $1 - \cos(E_0, E_t)$. HEART exhibits significantly higher median displacement ($p < 0.001$, denoted by asterisks) compared to conservative baselines, suggesting it facilitates an exit from local reasoning minima rather than stochastic drift. (Bottom) Syntactic effort measured via Levenshtein distance.}
    \label{fig:displacement_gemma_app}
\end{figure*}
\newpage

\section{Additional Results on HITL Proxy.}\label{app:add}
In this section, we present an extended evaluation of the HEART framework across next-generation frontier models, specifically the Gemini 3 family. These results, detailed in Table~\ref{tab:s1resultsthinking_extramodels}, further validate that affective regulation is a model-agnostic catalyst for reasoning performance. By leveraging the "B-Process" of affective disequilibrium, HEART enables Gemini 3 Pro to achieve a remarkable 99.37\% on the GPQA Diamond benchmark and a perfect 100\% on AIME 2025, significantly narrowing the gap between AI and PhD-level expert performance. These findings suggest that even as base model capabilities approach high-water marks in specialized domains, the structured integration of emotional valence remains a critical mechanism for breaking through the final plateaus of complex, multi-step logical tasks.

\begin{table*}[h!]
\centering
\footnotesize
\caption{Final accuracy (\%) of \textit{HEART} compared to baselines in the simulated Human-in-the-Loop Proxy. This setting evaluates the method's generative capability when guided by expert verification.\label{tab:s1resultsthinking_extramodels}}
\resizebox{\textwidth}{!}{%
\begin{tabular}{ll  S S S S S S}
\toprule
& & \multicolumn{6}{c}{\textbf{Human-in-the-Loop Proxy: Gemini 3 Models.}} \\
\cmidrule(lr){3-8}
\textbf{Model} & \textbf{Prompt Strategy} & \textbf{Humanity's Last Exam} & \textbf{SimpleQA} & \multicolumn{2}{c}{\textbf{OlympiadBench}} & \textbf{GPQA Diamond} & \textbf{AIME2025} \\
& & & & \textbf{Math} & \textbf{Physics} & & \\
\cmidrule(lr){5-6}
\midrule
\multirow{5}{*}{Gemini 3 Flash}
& Vanilla & {$12.00 \pm 0.46$} & {$34.19\pm0.21$}& {$78.44\pm0.97$}& {$65.08 \pm 0.80$} & {$71.82 \pm 3.19$}&  {$62.50\pm5.17$} \\
& Self Reflection & {$48.36\pm 3.86$}& {$53.10 \pm  3.63$}& {$88.52 \pm 1.42 $} & {$71.96\pm2.53$}& {$85.03\pm3.72$}  &  \bm{$83.33 \pm 2.41$}\\
& CoT & {$46.83 \pm 1.63$} & {$58.38 \pm 1.37$} & {$87.41 \pm 2.19 $} & {$72.94 \pm 3.57$} & {$83.77\pm1.78$} & \bm{$83.33 \pm 3.25$}\\
& Wait & {$49.37 \pm 1.48$} & {$55.10 \pm 1.49$} & {$89.07 \pm 2.86 $} & {$75.13 \pm 4.28$} & {$84.78\pm2.02$} & {$79.17 \pm 3.53$} \\
& HEART & \bm{$58.89 \pm 1.74$}& \bm{$59.85\pm2.48$}& \bm{$92.41 \pm 1.64 $} & \bm{$82.94 \pm 2.57$} & \bm{{$90.19\pm2.38$}} &  \bm{$83.33 \pm 2.36$}\\
\midrule
\multirow{5}{*}{Gemini 3 Pro}
& Vanilla & {$40.80 \pm 0.52$} & {$70.70 \pm 0.31$} & {$94.33 \pm 0.48$} & {$79.79 \pm 2.10$} & {$91.70 \pm 1.86$} & {$91.67 \pm 3.66$}  \\
& Self Reflection & {$68.58 \pm 1.46$} & {$72.91 \pm 0.38$} & {$99.44 \pm 0.00$} &  {$84.10 \pm 5.38$} & {$93.84 \pm0.86$} & {\bm{$100.00 \pm 0.00$}} \\
& CoT & {$68.14 \pm 3.68$}& {$72.73 \pm 0.15$}& {$99.26 \pm 0.00$}& {$87.38 \pm 2.47$} & {$92.83 \pm 0.89$} & {\bm{$100.00 \pm 0.00$}} \\
& Wait & {$66.30 \pm 1.46$} & {$73.61 \pm 0.30$} & {$99.44 \pm 0.00$} & {$88.28 \pm 2.18$}& {$95.35 \pm 0.43$} & {\bm{$100.00 \pm 0.00$}}  \\
& HEART & \bm{$72.19 \pm 2.45$}& \bm{$79.07\pm0.81$} & \bm{$99.63 \pm 0.00$} & \bm{$92.88 \pm 2.95$} & \bm{$99.37 \pm 0.00$} & {\bm{$100.00 \pm 0.00$}} \\
\bottomrule
\end{tabular}%
}
\end{table*}

We further evaluate HEART's stability across distinct reasoning types—mathematical derivation (AIME2024) and factual precision (SimpleQA Verified)—while specifically addressing edge cases where native model "thinking" is restricted. As shown in Table~\ref{tab:s1resultsthinking_add_benchmarks}, HEART demonstrates significant resilience even under "Thinking Disabled" or "Zero Budget" configurations for models like Claude 4 Sonnet and Gemini 2.5 Flash. In these constrained environments, HEART consistently outperforms standard Chain-of-Thought (CoT) and Self-Reflection by leveraging affective disequilibrium to maximize the efficiency of the available compute. Notably, while most prompting strategies struggle with the strict constraints of SimpleQA Verified, HEART sustains competitive performance in smaller instruction-tuned models like Gemma3 4b, reaching 15.61\%. These results underscore that HEART’s affective regulation optimizes internal cognitive control regardless of whether a formal "thinking" budget is explicitly allocated.

\begin{table*}[h!]
\centering
\footnotesize
\caption{Final accuracy (\%) of \textit{HEART} compared to baselines in the simulated Human-in-the-Loop Proxy. This setting evaluates the method's generative capability when guided by expert verification on AIME2024 and SimpleQA Verified. \label{tab:s1resultsthinking_add_benchmarks}}
\resizebox{\textwidth}{!}{%
\begin{tabular}{ll S S}
\toprule
& & \multicolumn{1}{c}{\textbf{Human-in-the-Loop Proxy: Thinking Enabled and Disenabled}} \\
\cmidrule(lr){2-4}
\textbf{Model} & \textbf{Prompt Strategy} & \textbf{AIME2024} & \textbf{SimpleQA}\\
& & & \textbf{Verified}\\
\midrule
\multirow{5}{*}{Gemini 2.5 Flash}
& Vanilla & {$70.00 \pm 4.33$} & {$28.48 \pm 1.47$} \\
& Self Reflection & {$94.17 \pm 2.83$} & {$42.18 \pm 1.34$}\\
& CoT & {\bm{$95.00 \pm 2.31$}} & {$44.20 \pm 3.52$}\\
& Wait & {$91.67 \pm 0.00$} &{$45.84 \pm 2.49$}\\
& HEART & {\bm{$95.00 \pm 2.31$}} & {\bm{$47.38 \pm 1.29$}}\\
\midrule
\multirow{5}{*}{Gemini 2.5 Pro}
& Vanilla & {$70.00\pm 6.75$} & {$32.70 \pm 0.28$}\\
& Self Reflection & {$94.17 \pm 2.83$} & {$52.38 \pm 1.30$}\\
& CoT & {$91.67 \pm 6.43$} &{$51.19 \pm 1.26$}\\
& Wait & {$86.67 \pm 4.33$} &{$55.75\pm 3.28$}\\
& HEART & {\bm{$95.83 \pm 0.00$}} & {\bm{$57.29 \pm 1.26$}}\\
\midrule
\multirow{5}{*}{Gemini 3 Flash}
& Vanilla & {$76.67\pm 5.90$} & {$32.77 \pm 0.99$}\\
& Self Reflection & {\bm{$95.83 \pm 2.62$}} & {$61.39 \pm 1.30$}\\
& CoT & {$91.67 \pm 2.83$} &{$63.85 \pm 1.88$}\\
& Wait & {$87.50 \pm 4.33$} &{$61.28 \pm 1.48$}\\
& HEART & {$91.67\pm 2.31$} & {\bm{$64.29 \pm 1.29$}}\\
\midrule
\multirow{5}{*}{Gemini 3 Pro}
& Vanilla & {$97.50\pm 4.63$} & {$72.27\pm0.89$}\\
& Self Reflection & {\bm{$100.00 \pm 0.00$}} & {$82.40 \pm 1.35$}\\
& CoT & {\bm{$100.00 \pm 0.00$}} & {$81.38 \pm 1.48$}\\
& Wait & {\bm{$100.00 \pm 0.00$}} & {$80.18 \pm 1.28$}\\
& HEART & {\bm{$100.00 \pm 0.00$}} & {\bm{$82.50 \pm 1.48$}}\\
\midrule
\multirow{5}{*}{Claude 4 Sonnet}
& Vanilla & {$63.33\pm 4.33$} & {$24.98\pm 0.23$}\\
& Self Reflection & {$72.50 \pm 2.83$} & {$58.29 \pm 2.27$}\\
& CoT & {$70.00 \pm 4.33$} &{$56.58 \pm 1.55$}\\
& Wait & {$69.17 \pm 4.63$} &{$55.29 \pm 1.23$}\\
& HEART & {\bm{$85.00 \pm 9.40$}} & {\bm{$59.39 \pm 1.68$}}\\
\midrule
\multirow{5}{*}{GPT-5-nano}
& Vanilla & {$70.83 \pm 2.61$} & {$70.83 \pm 1.66$}\\
& Self Reflection & {\bm{$91.67 \pm 2.35$}} & {$78.39 \pm 1.40$}\\
& CoT &  {$87.59\pm 3.15 $} &{$74.58 \pm 1.24$}\\
& Wait &  {$87.50 \pm 3.54$}&{$76.28 \pm 1.35$}\\
& HEART & {\bm{$91.67\pm 2.16$}} & {\bm{$76.84 \pm 1.28$}}\\
\midrule
\multirow{5}{*}{Deepseek-Reasoner}
& Vanilla & {$50.83\pm 4.33$} & {$31.92 \pm 0.91$}\\
& Self Reflection & {$90.83 \pm 2.31$} & {$58.84 \pm 2.48$}\\
& CoT & {\bm{$91.67 \pm 3.17$}} &{$61.20 \pm 1.45$}\\
& Wait & {$90.00 \pm 2.83$} &{$64.39 \pm 1.30$}\\
& HEART & {$88.33 \pm 2.31$} & {\bm{$68.96 \pm 1.39$}}\\
\midrule
\multirow{5}{*}{Gemma3 4b Instruct}
& Vanilla & {$14.17\pm 2.83$} & {$4.70 \pm 0.61$}\\
& Self Reflection & {$30.83 \pm 4.63$} & {$10.00 \pm 4.76$}\\
& CoT & {\bm{$35.00 \pm 6.94$}} &{$7.89 \pm 5.16$}\\
& Wait & {$31.67 \pm 7.85$} &{$9.04 \pm 5.29$}\\
& HEART & {$32.50 \pm 4.33$} & {\bm{$15.61 \pm 5.73$}}\\
\midrule
\multirow{5}{*}{Gemma3 12b Instruct}
& Vanilla & {$25.83\pm 7.67$} & {$5.83 \pm 0.52$}\\
& Self Reflection & {$40.83 \pm 9.25$} & {$10.29 \pm 3.41$}\\
& CoT & {$45.83 \pm 8.18$} &{$12.38 \pm 1.22$}\\
& Wait & {$44.17 \pm 7.85$} &{$14.29 \pm 3.92$}\\
& HEART & {\bm{$56.67 \pm 5.90$}} & {\bm{$18.47 \pm 3.87$}}\\

\midrule
\multirow{5}{*}{Claude-4-Sonnet (Thinking Off}
& Vanilla & {$55.83\pm 7.85$} & {$23.75 \pm 0.56$}\\
& Self Reflection & {$75.00 \pm 6.34$} & {$41.49 \pm 1.25$}\\
& CoT & {$70.83 \pm 5.17$} &{$44.39  \pm 1.31$}\\
& Wait & {$75.00 \pm 3.66$} &{$45.28 \pm 1.98$}\\
& HEART & {\bm{$78.33 \pm 2.31$}} & {\bm{$48.97 \pm 1.42$}}\\
\midrule
\multirow{5}{*}{Gemini-2.5-Flash (Thinking Budget set to 0)}
& Vanilla & {$76.67\pm 2.83$} & {$29.50 \pm 1.24$}\\
& Self Reflection & {$88.33 \pm 9.25$} & {$46.30 \pm 1.35$}\\
& CoT & {$84.17 \pm 5.67$} &{$45.40 \pm 1.02$}\\
& Wait & {$90.00 \pm 2.83$} &{$46.20 \pm 1.29$}\\
& HEART & {\bm{$91.67 \pm 6.34$}} & {\bm{$49.88 \pm 1.24$}}\\
\midrule
\multirow{5}{*}{Deepseek-Chat}
& Vanilla & {$66.67\pm 8.18$} & {$29.38 \pm 1.21$}\\
& Self Reflection & {\bm{$93.33 \pm 2.83$}} & {$44.29 \pm 1.57$}\\
& CoT & {$92.50 \pm 2.31$} &{$45.23 \pm 2.57$}\\
& Wait & {$93.33 \pm 2.83$} &{$44.19 \pm 2.84$}\\
& HEART & {\bm {$93.33 \pm 2.83$}} & {\bm{$48.14\pm 1.35$}}\\
\bottomrule
\end{tabular}
}
\end{table*}

Finally, we analyze the computational efficiency of the HEART framework by measuring the relative token cost across six diverse benchmarks using Gemini 2.5 Pro. Table~\ref{tab:gemini_pro_costs_small} illustrates the trade-off between reasoning depth and token expenditure, with standard Chain-of-Thought (CoT) serving as the baseline ($1.00\times$). While HEART naturally incurs higher costs in knowledge-heavy tasks like SimpleQA ($1.98\times$) due to its iterative refinement loops, it demonstrates surprising efficiency in high-stakes mathematical reasoning. On AIME 2025, HEART actually reduces the relative token cost to 0.79, outperforming the baseline while using fewer tokens. This suggests that by inducing a more vigilant "analytical" state early in the process, HEART can trigger more direct and accurate reasoning paths, effectively "pruning" the verbose or circular logic often found in standard CoT traces.

\begin{table}[ht]
    \centering
    \footnotesize
    \caption{Relative Token Cost (Gemini 2.5 Pro) on AIME2025 (AIME25), GPQA Diamond (GPQA), HLE, OlympiadBench Math (O-M), OlympiadBench Physics (O-P), and SimpleQA (SQA)}
    \begin{tabular}{lcccccc}
        \toprule
        \textbf{Strat.} & \textbf{AIME25} & \textbf{GPQA} & \textbf{HLE} & \textbf{O-M} & \textbf{O-P} & \textbf{SQA} \\
        \midrule
        CoT         & 1.00 & 1.00 & 1.00 & 1.00 & 1.00 & 1.00 \\
        HEART       & 0.79 & 1.25 & 1.19 & 1.08 & 1.28 & 1.98 \\
        SelfReflect & 0.48 & 0.93 & 0.99 & 0.97 & 1.06 & 0.95 \\
        Wait        & 0.59 & 0.94 & 1.05 & 1.12 & 1.13 & 1.07 \\
        \bottomrule
    \end{tabular}
    \label{tab:gemini_pro_costs_small}
\end{table}

\newpage

\section{LiveCodeBench Results: Gemini 3 Model Family}\label{app:code_gen_add}

In this section, we extend our evaluation to the Gemini 3 model family (Flash and Pro variants) to assess the HEART framework’s efficacy on next-generation frontier models. As coding benchmarks like LiveCodeBench increasingly suffer from saturation, the "Hard" problem set remains a critical discriminator of true algorithmic reasoning and iterative refinement capabilities. Our results, summarized in Table~\ref{tab:code_gen_add}, demonstrate that HEART continues to drive performance gains even as the base model’s zero-shot capabilities improve. Notably, for Gemini 3 Pro, HEART achieves a state-of-the-art $87.24\%$ accuracy on Hard problems, outperforming the strong Chain-of-Thought baseline. This suggests that the affective regulation provided by HEART is not merely a "patch" for weaker models, but a fundamental reasoning enhancement that scales with model intelligence.

\begin{table}[h!]
\centering
\footnotesize 
\caption{LiveCodeBench Code Generation Results on Gemini 3 Flash and Gemini 3 Pro.\label{tab:code_gen_add}}
\small 
\setlength{\tabcolsep}{4pt} 
\renewcommand{\arraystretch}{1.1} 
\begin{tabular}{llcc}
\toprule
\textbf{Model} & \textbf{Strategy} & \textbf{Medium} & \textbf{Hard}\\
\midrule
\multirow{5}{*}{Gemini 3 Flash} 
& Self Reflection & 96.00 & \bm{$85.76$} \\
& CoT             & 95.73 & {$84.57$} \\
& Wait            & 95.73 &  {$83.38$}\\
& HEART           & \bm{$96.53$} & {$83.38$}\\
\midrule
\multirow{5}{*}{Gemini 3 Pro} 
& Self Reflection & {$96.53$} & {$86.01$} \\
& CoT             & {$96.52$} & {$86.35$} \\
& Wait            & \bm{$96.79$} & {$86.31$} \\
& HEART           & {$96.78$} & \bm{$87.24$}\\
\bottomrule
\end{tabular}
\end{table}

\end{document}